\documentclass[dvipsnames,format=sigconf,anonymous=false,review=false]{acmart}

\usepackage{hyperref}
\usepackage{multirow}
\usepackage{xcolor,colortbl}
\usepackage{graphicx}
\usepackage{subcaption}
\usepackage{balance}

\newcommand{\best}[1]{\cellcolor{gray}\textcolor{white}{\textbf{{#1}}}}

\AtBeginDocument{%
  \providecommand\BibTeX{{%
    \normalfont B\kern-0.5em{\scshape i\kern-0.25em b}\kern-0.8em\TeX}}}

\begin{document}

\title{A Satellite Band Selection Framework for Amazon Forest Deforestation Detection Task}

\author{Eduardo Neto and Fabio A. Faria}
\authornote{Both authors contributed equally to this research.}
\affiliation{\institution{Institute of Science and Technology\\Universidade Federal de S\~ao Paulo}
  \city{S\~ao Jos\'e dos Campos}
  \state{SP}
  \country{Brazil}
} 
\email{{ebneto,ffaria}@unifesp.br}

\author{Amanda A. S. de Oliveira}
\affiliation{\institution{Escola de Artes, Ciências e Humanidades \\University of Sao Paulo}
  \city{S\~ao Paulo}
  \state{SP}
  \country{Brazil}
}
\email{amandaoli@alumni.usp.br}

\author{Álvaro Fazenda}

\affiliation{%
  \institution{Institute of Science and Technology\\Universidade Federal de S\~ao Paulo}
  \city{S\~ao Jos\'e dos Campos}
  \state{SP}
  \country{Brazil}
}\email{alvaro.fazenda@unifesp.br }


\begin{abstract}
 The conservation of tropical forests is a topic of significant  social and ecological relevance due to their crucial role in the global ecosystem. Unfortunately, deforestation and degradation impact millions of hectares annually, necessitating government or private initiatives for effective forest monitoring.
  This study introduces a novel framework that employs the Univariate Marginal Distribution Algorithm (UMDA) to select spectral bands from Landsat-8 satellite, optimizing the representation of deforested areas.  This selection guides a semantic segmentation architecture, DeepLabv3$+$, enhancing its performance. Experimental results revealed several band compositions that achieved superior balanced accuracy compared to commonly adopted combinations for deforestation detection, utilizing segment classification via a Support Vector Machine (SVM). Moreover, the optimal band compositions identified by the UMDA-based approach improved the performance of the DeepLabv3$+$ architecture, surpassing state-of-the-art approaches compared in this study. The observation that a few selected bands outperform the total contradicts the data-driven paradigm prevalent in the deep learning field. Therefore, this suggests an exception to the conventional wisdom that ‘more is always better’.

 
\end{abstract}



\maketitle

\section{Introduction}
Tropical forests are located between the Tropics of Cancer and Capricorn, near the Equator, and are found in South and Central America, Africa, and regions of Asia and the Pacific. 
Taking into account the Amazon Forest, which is the largest tropical forest, it is estimated that approximately 38.0\% of the remaining forests in the region are currently degraded by fire, edge effects, timber extraction, and/or extreme drought~\cite{doi:10.1126/science.abp8622}. 
In addition, these conditions lead to a reduction in dry-season evapotranspiration by up to 34.0\%, resulting in biodiversity loss due to deforestation in human-altered landscapes. This, in turn, generates uneven socioeconomic burdens, predominantly affecting forest dwellers.
Despite tropical forests constituting a mere 7.0\% of the Earth’s surface, they are estimated to be home to over half of the planet’s species~\cite{martin2015edge}. 
In addition to the high biodiversity, the plants and soil of tropical forests store between $460$ and $575$ billion tonnes of carbon. The processes of evaporation and evapotranspiration from their plants and trees release large amounts of water into the local atmosphere, promoting the formation of clouds and rainfall~\cite{nasaflorestas}.

Unfortunately, millions of acres of tropical rainforests are degraded every year through deforestation~\cite{martin2015edge, Hansen850}. According to the well-known PRODES (Brazilian Legal Amazon Monitoring Program by Satellite), in the timespan betweeen August/2021 and July/2022, over $11.000$ $km^2$ of forest were degraded~\cite{prodes2022}. To make matters worse, due to the shortage of qualified professionals and the large amounts of data that need to be analyzed, forest monitoring programs such as PRODES are very expensive, both in terms of time and resources.

Machine learning techniques and digital image processing have been developed to overcome such challenges and assist experts in tasks related to automatic or semi-automatic detection of deforested areas in tropical forests. Ortega {et al.}~\cite{ortega2019evaluation} affirms that the use of classification methods based on deep learning — particularly Siamese Convolutional Neural Networks and {Early Fusion} methods – is a superior approach to traditional Machine Learning techniques. Andrade {et al.}~\cite{andrade2020evaluation} addresses the deforestation detection task in satellite images through semantic segmentation (SS) using the DeepLabv3$+$ model. Both works utilize Landsat-8 satellite images with all seven spectral bands in addition to Normalized Difference Vegetation Index (NDVI). In Maretto {et al.}~\cite{maretto2020spatio}, an approach for deforestation detection via semantic segmentation of Landsat-8 images with image fusion based on the well-known U-Net architecture was proposed. For this purpose, images composed of 5 bands (B3, B4, B5, B6, and B7) were chosen {ad-hoc}.

In order to explore the potential of multispectral sensors (such as Landsat-8) to its fullest, many studies have been proposed to create new images of the same geographical area by selecting spectral bands and seeking the best representations for the target application. Yu {et al.}~\cite{yu2019selection} proposes a Landsat-8 band selection methodology based on their correlations to optimize the performance of a Support Vector Machine (SVM) in land use/land cover classification. On the other hand, Dallaqua et al.~\cite{foresteyes2019} conducted a "brute force" study in the search for the best band composition that would provide higher accuracy in deforestation detection by an SVM classifier. In this context, the use of search algorithms proves to be a good alternative solution, as it is more efficient in terms of time and computational cost than traditional methods such as conventional grid search~\cite{ma2003bandselection}. Also in the scope of Remote Sensing (RS) imagery, Zhang {et al.}~\cite{zhang2009bandselection} uses a genetic coding approach to perform band selection in high dimensional RS images. The chosen fitness function was based on a distance index that indicates the separability between the classes of interest.

In the agronomy field, Nagasubramanian {et al.}~\cite{nagasubramanian2018hyperspectral_ga} used a genetic algorithm in the search for the best band combination of a hyperspectral sensor for pest detection in soybean plantations. Through a genetic algorithm, they determine the optimal spectral combination for disease identification using hyperspectral images of plant stems. The fitness function (which guides the genetic algorithm) is based on the classification performance of a Support Vector Machine. The results obtained with the band combination identified by the genetic algorithm show a classification accuracy improvement of $26.1\%$ compared to conventional RGB images. A ``brute force" study was also conducted, yielding the same optimal combination as the proposed feature selection algorithm.

The combined use of Haralick descriptors and evolutionary algorithms for feature selection was employed in~\cite{amroabadi2011haralickmammograms} for tumor detection in mammography images. The features under consideration are the vectors produced by the Principal Component Analysis (PCA) and the Haralick texture descriptors~\cite{HARALICK} of the images and the fitness function implementation is based on a neural network classifier. In other words, the combinations of vectors and texture descriptors that yield the best performance metrics in the classifier are selected. The results indicate that the use of feature selection led to an enhancement of 4.0\% in the area under curve (ROC).

Feature selection algorithms can also be used for tasks outside the image processing domain. Wei et al.~\cite{WeiFeatureSelection2017} uses the Univariate Marginal Distribution Algorithm (UMDA) to determine the best set of features for four different classification datasets in different domains ranging from bioinformatics to audio processing. A Support Vector Machine was used as the classifier. By combining Rough Set Theory with UMDA, an improvement of up to 6.0\% in classification accuracy can be achieved, reducing data dimension and accelerating classifier convergence.


In the field of bioinformatics, Saeys {et al.}~\cite{saeys2004featureRanking} proposed a method based on an Estimation of Distribution Algorithm (EDA) to select the most relevant nucleotides in a splice site prediction task. In this work, the authors not only reached the solution of the search problem through the best individual, but also inferred a feature ranking. Thus, each feature in the final population has its probability estimated. Features with higher probabilities are assumed to be more relevant to the task at hand than the ones with lower probabilities. The results showed that, up to a certain limit, the removal of features does not decrease the performance of the splice site predictor significantly. In some of the datasets used for the study, it was shown that discarding the lower-ranking features might actually increase the prediction performance.

Therefore, in the sense to enhance the efficacy of deforestation detection models, we introduce a comprehensive band selection framework founded on the Univariate Marginal Distribution Algorithm (UMDA). This novel approach aims to systematically identify and prioritize spectral bands from multispectral data, optimizing the input features for subsequent analysis. Particularly, in this work, we use a semantic segmentation model for the deforestation detection task.

\section{Fundamental Concepts}
In this section, we delve into the fundamental concepts underpinning our study, providing a solid foundation for the subsequent analyses.

\subsection{Landsat Satellite} 

The Landsat series began to be conceived in the 1960s, when there was pressure from both the scientific community, the government and the private sector, to create a satellite program for observing Earth's resources. NASA, in cooperation with DOI ({U.S. Department of the Interior}), USDA ({United States Department of Agriculture}), and other agencies, obtained resources for an Earth observation satellite, which was launched successfully in $1972$, called {Earth Resources Technology Satellite} (ERTS-1), which was later called Landsat-1

In $2013$, Landsat-8 satellite was launched, developed in collaboration between NASA and the USGS ({U. S. Geological Survey})~\cite{landsat8Specs}. It is located at an altitude of $705km$ from Earth, with a revisit rate of $16$ days and $185km$ of imaged swath on Earth. The satellite has two scientific instruments: {Operational Land Imager} (OLI) and {Thermal Infrared Sensor} (TIRS). Measures $11$ different frequency bands along the electromagnetic spectrum (color)~\cite{landsat8Overview}. Each band is called a spectral band.  Table \ref{tab:land8} presents the characteristics of these bands.

\begin{table}[ht!]
\centering
\caption{The band designations for the Landsat-8 satellite.}
\label{tab:land8}
  \resizebox{8.50cm}{!}{
  
\begin{tabular}{@{}clcc@{}}
\toprule
\textbf{Sensor}              & \multicolumn{1}{c}{\textbf{Bands}} & \begin{tabular}[c]{@{}c@{}}\textbf{Wavelength}\\ \textbf{(micrometers)}\end{tabular} & \begin{tabular}[c]{@{}c@{}}\textbf{Resolution}\\ \textbf{(meters)}\end{tabular} \\ \midrule
\multirow{9}{*}{OLI}  
&  (B1) Coastal aerosol  & 0.433 - 0.453     & 30          \\
& (B2) Blue   & 0.450 - 0.515     & 30          \\
& (B3) Green  & 0.525 - 0.600     & 30          \\
& (B4) Red     & 0.630 - 0.680     & 30          \\
& (B5) Near InfraRed  & 0.845 - 0.885     & 30          \\
& (B6) Shortwave  InfraRed 1   & 1.560 - 1.660     & 30          \\
& (B7) Shortwave  InfraRed 2   & 2.100 - 2.300     & 30          \\
& (B8) Pancromático & 0.500 - 0.680     & 15          \\
& (B9) Cirrus & 1.360 - 1.390     & 30          \\ \midrule
\multirow{2}{*}{TIRS} & (B10) InfraRed  Termal & 10.6 - 11.2       & 100         \\
& (B11) InfraRed Termal& 11.5 - 12.5       & 100         \\ \bottomrule 
\end{tabular}
}
\end{table}

\subsection{PRODES}

Amazon Deforestation Monitoring Project (PRODES)~\cite{AFW14,metodprodes2019} was developed by the National Institute for Space Research (INPE, in the Portuguese acronym) in $1988$ to carry annual deforestation surveys in BLA. Typically, PRODES reports are published between July and August. Therefore, images from this time period are used to track yearly deforestation. Trained specialists classify remote sensing images through photointerpretation. These remote sensing images are usually from Landsat satellites, but other satellites as SENTINEL-2 and CBERS-4 can also be used. In $2013$, Landsat-8~\cite{landsat8Specs} was launched, which has a revisit time of $16$ days and captures images with up to $10$ meters of spatial resolution in $11$ spectral bands. However, following other Landsat-based automatic deforestation works~\cite{torres_2021, andrade2020evaluation, ortega2019evaluation}, only bands B1 through B7 are considered in this study.

\begin{figure*}[!ht]
\centering
\begin{tabular}{cc}
\includegraphics[height=0.3\textwidth,keepaspectratio=true]{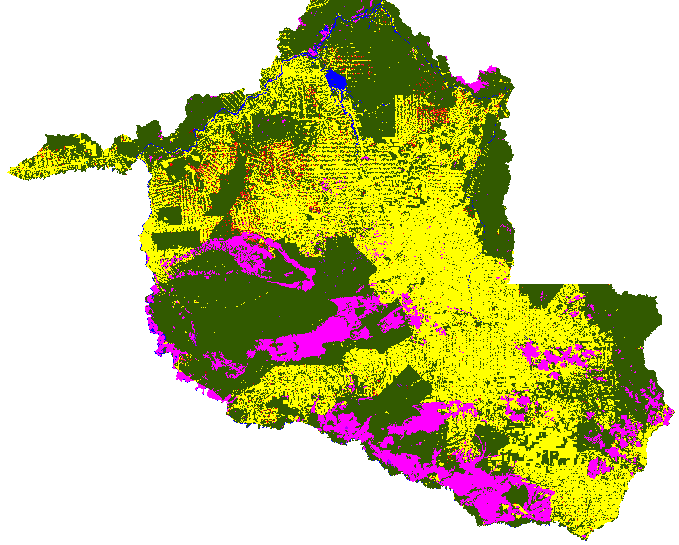} & 
\includegraphics[height=0.3\textwidth,keepaspectratio=true]{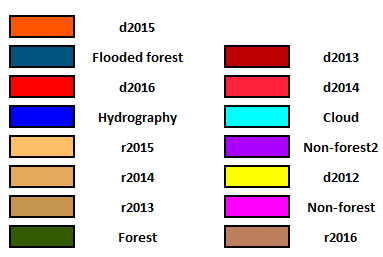} \\
(a) PRODES' mosaic of the State of Rondônia for the year of $2016$. &  (b) Different classes existing on PRODES $2016$. \\
\end{tabular}
\caption{Rondônia's thematic image for the PRODES year of $2016$. Extracted from TerraBrasilis~\cite{deterrabrasilis}. In (a) Thematic image of Rondônia in the year $2016$ and in 
(b) Legend for the thematic image where "d" means deforestation of the given year and "r" is the residue. The residue is deforestation that was not detected but occurred in previous years~\cite{macielreasoning}.} 
\label{fig:imagensPRODES2016}
\end{figure*}

PRODES methodology is divided into three steps: (1) an image selection technique is performed, where cloud-free images collected near August 1st are enhanced to highlight the areas with complete removal of native forest (clear-cutting); (2) the specialists identify and delimit the deforestation polygons; and (3) the annual deforestation rate is calculated~\cite{metodprodes2019}. Both thematic images and quantitative data are publicly available. Figure~\ref{fig:imagensPRODES2016} presents the PRODES' mosaic for the State of Rondônia for the year $2016$.

As PRODES quantifies and spatially locates annual BLA deforestation increments, once an area is classified as deforestation, it is included in an exclusion mask that will be used in subsequent years. Therefore, PRODES data don't infer on regenerated areas~\cite{PRODES3}. This exclusion mask also has regions of non-original vegetation (which PRODES classifies as non-forest) and hydrography areas.

\subsection{Superpixels with SLIC}

Superpixels approaches, defined as clusters of connected pixels with similar characteristics, are used in numerous applications necessitating image segmentation. By reducing the pixel count in an image, superpixels enhance computational efficiency and reducing processing time. The similarity within a superpixel is quantified using a measure derived from attributes such as intensity, color, texture, and position~\cite{alexandre2017ift}.

The Simple Linear Iterative Clustering (SLIC) algorithm employs K-means clustering to extract superpixels in color images considering the CIELAB color space, thereby generating the Labxy feature space. It takes parameters such as the desired number $k$ of superpixels and control over their compactness $m$. The latter ensures that superpixels exhibit uniform sizes and shapes, where lower values result in superpixels with irregular sizes and shapes~\cite{SLIC, alexandre2017ift}.

A significant advantage of the SLIC algorithm is its linear complexity relative to the number $N$ of pixels in the input image, as its search space is limited to the desired size of the superpixel.

\subsection{\textbf{Semantic Segmentation with DeepLabv3$+$}}
Semantic segmentation task consists in the assignment of class labels to individual pixels within an image~\cite{chen2023generative, chen2018encoderdecoder}. As stated by Long et al.~\cite{long2015fully}, this task represents a logical progression from coarse to refined inference. The outcome of a semantic segmentation model is an image that retains the original input dimensions, with each pixel assigned a specific class. Notably, two or more instances of the same class are represented with the same color, that is, there is no distinction between occurrences of a class. This proves suitable for the task of deforestation detection, as our focus lies not necessarily on individual deforestation spots, but rather on the assessment of the entire deforested area.

In this paper, to ensure a fair comparison with the selected semantic segmentation baseline approach~\cite{torres_2021}, we also employed the widely recognized DeepLabv3$+$ architecture  proposed by Chen et al.~\cite{chen2018encoderdecoder}. This Atrous convolution-based architecture is an extended version of the DeepLabv3 that follows an encoder-decoder paradigm, where the encoder is based on a pre-trained ResNet-101 from Imagenet~\cite{deng2009imagenet}. DeepLabv3$+$ adds a simple and effective decoder module to refine segmentation results, especially along object boundaries.

\subsection{Univariate Marginal Distribution (UMDA)}
\label{sec:umda}
Introduced by Mühlenbein \& Paa\ss~\cite{muhlenbein1996recombination}, the {Univariate Marginal Distribution} algorithm (UMDA) is one of the simplest Estimation of Distribution Algorithms~\cite{HAUSCHILD2011111} (EDAs). To optimize a pseudo-Boolean function $f: \{0,1\}^n \rightarrow \mathbb{R}$ where an individual is a bit-string (each gene is $0$ or $1$), the algorithm follows an iterative process: 1) independently and identically sampling a population of $\lambda$ individuals (solutions) from the current probabilistic model; 2) evaluating the solutions; 3) updating the model from the fittest $\mu$ solutions. Each sampling-update cycle is called a generation or iteration. In each iteration, the probabilistic model in the generation $t \in \mathbb{N}$ is represented as a vector $p_t=(p_t(1), \dots ,p_t(n))\in [0,1]^n$, where each component (or marginal) $p_t(i)\in[0,1]$ to $i\in[n]$, and $t \in \mathbb{N}$ is the probability of sampling the number one in the $i^{th}$ position of an individual in the generation $t$. Each individual $x = (x_1,  \dots, x_n) \in \{0, 1\}^n$ is therefore sampled from the joint probability.

\begin{equation}
    Pr(x | p_t) = \prod_{i=1}^{n}p_t(i)^{x_i}(1-p_t(i))^{(1-{x_i})}.
    \label{equacaoPr}
\end{equation}

Extreme probabilities, zero and one, should be avoided for each marginal point $(i)$; otherwise, the bit at position $i$ would never have its value changed, making the optimization ignore some regions of the search space. To avoid this, all $p_{t + 1} (i)$ margins are usually restricted within the closed range $[1/n, 1-1/n]$ and these values $1/n$ and $1-1/n$ are called lower and upper bounds respectively. In this case, UMDA is known as "UMDA with margins".

\textcolor{red}{Melhorar explicação do UMDA}
\subsection{Homogeneity Rate (\textit{HoR})}
To quantify the quality of image segments produced by an algorithm such as SLIC~\cite{SLIC}, which does not assign classes to the generated segments,~\cite{dallaquaSIBGRAPI} proposes the use of the Homogeneity Ratio (or \textit{HoR}). It is calculated based on the ground truth of the segmented region. The ratio is defined as the number of pixels belonging to the majority class (according to the ground truth) divided by the total number of pixels in the segment. Equation \eqref{eq:hor} formalizes this definition applied to the context of the present work, where binary segmentation is considered (classes of "forest" and "non-forest"). In this equation, ``NFP" refers to the number of pixels in the "forest" class, while ``NNP" refers to the number of pixels in the "non-forest" class.

\begin{equation}
\label{eq:hor}
HoR = \frac{max(NFP,NNP)}{NFP + NNP}
\end{equation}

\subsection{Evaluation Metrics}
Evaluation metrics serve as the benchmark for performance assessment in specified tasks. Within the scope of this study, the tasks under consideration are binary classification and binary semantic segmentation. For both tasks, deforestation (alternatively referred to as ``non-forest") is deemed as the positive class, signifying it as the class of interest. Consequently, the evaluation metrics employed in this context are delineated below.

\subsubsection{\textbf{Precision:}}
measures the accuracy of the positive predictions made by a model. That is, a higher precision value means fewer occurrences of false positive.
\begin{equation}
    \text{Precision} = \frac{\text{True Positive}}{\text{True Positive} + \text{True Positive}} \label{eq:precision}
\end{equation}

\subsubsection{\textbf{Recall:}}
 assesses the ability of a model to detect all instances of the positive class across a dataset. A higher recall indicates fewer false negative.
\begin{equation}
    \text{Recall} = \frac{\text{True Positive}}{\text{True Positive} + \text{False Negative}} \label{eq:recall}
\end{equation}

\subsubsection{\textbf{F1-Score:}}
 consists in the harmonic mean between Precision and Recall, providing a balance between both metrics.
\begin{equation}
    \text{F1-Score} = \frac{2 \times \text{Precision} \times \text{Recall}}{\text{Precision} + \text{Recall}} \label{eq:f1score}
\end{equation}

\subsubsection{\textbf{Accuracy:}}
 provides a measure of the overall correctness of the model predictions.
\begin{equation}
    \small \text{Acc} = \frac{\text{True Positive} + \text{True Negative}}{\text{True Positive} + \text{True Positive} + \text{True Negative} + \text{False Negative}} \label{eq:accuracy}
\end{equation}

\subsubsection{\textbf{Intersection over Union (IoU):}}
When it comes to semantic segmentation, traditional performance metrics can be misleading. This occurs due to the complex nature of object boundaries and intricacies such as partial occlusion and variance in scale between classes. Additionally, the presence of class imbalance introduces a potential bias in the evaluation metrics. To handle such challenges, the Intersection over Union (IoU) index is one of the standard indicators for evaluating SS performance~\cite{tian2023AtmosphericRivers}. It is formally defined as the overlap ratio between the ground-truth and the model output~\cite{Nowozin_2014_CVPR}. The IoU index can also be derived from the model predictions at a pixel level by using Equation ~\ref{eq:iou}~\cite{Fernandez2018SSmetrics}.

\begin{equation}
    \text{IoU} = \frac{\text{True Positive}}{\text{True Positive} + \text{True Negative} + \text{True Positive}} \label{eq:iou}
\end{equation}

\section{The Landsat Band Selection Framework}
\label{sec:fundamentals}
\begin{figure*}[ht!]
    \centering
    \includegraphics[width=0.8\linewidth]{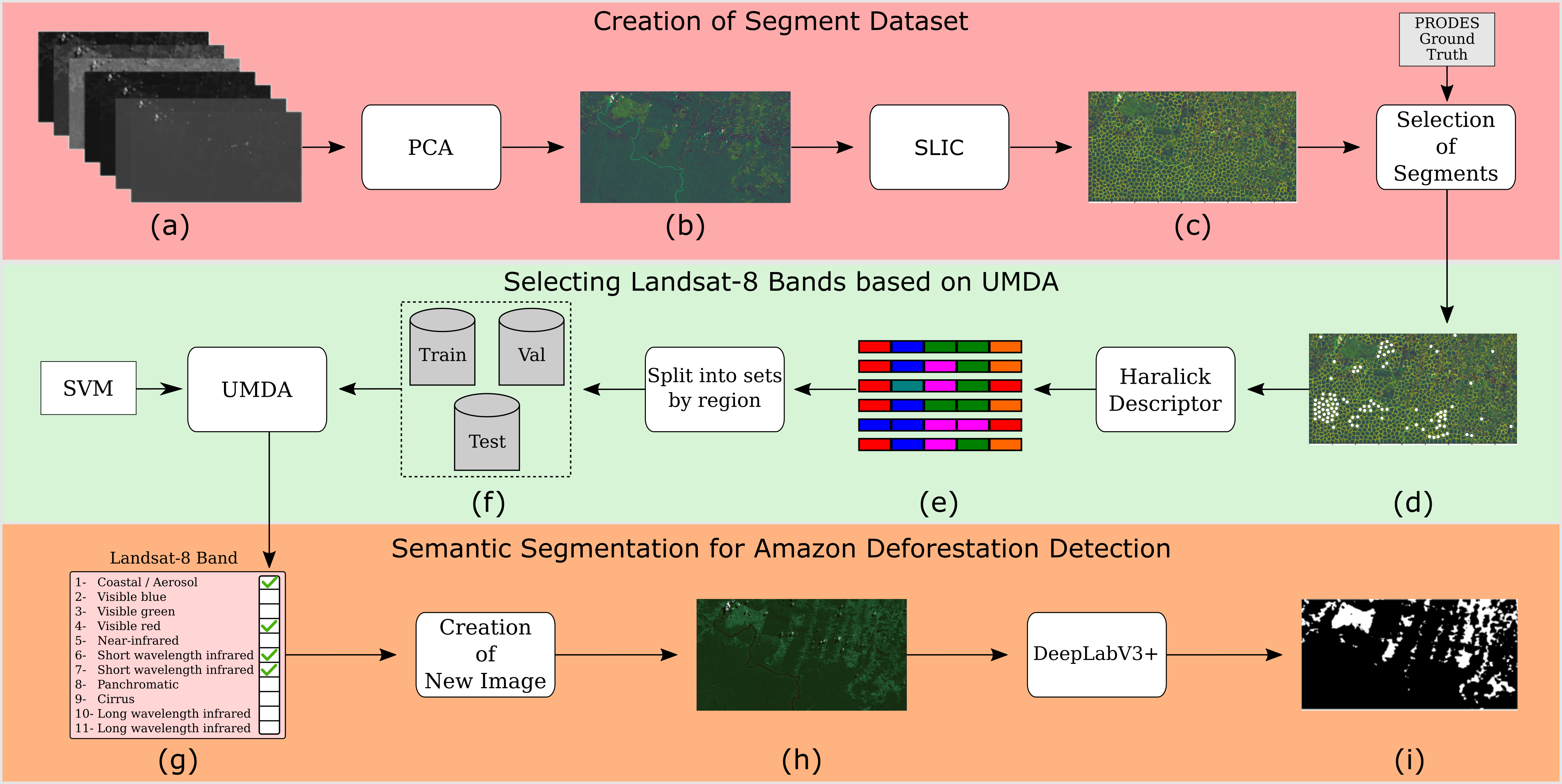}
    \caption{The pipeline of the Landsat-8 band selection framework based on UMDA for deforestation detection is depicted as follows: (a) displays an image from the Landsat-8 imaging satellite with its seven spectral bands, (b) represents the false-color image formed by three principal components of PCA~\cite{PCA}, (c) illustrates the output of the SLIC~\cite{SLIC} superpixel generation with various identified segments, (d) showcases the resulting image from the selection of segments indicated by white dots, (e) illustrates the Haralick~\cite{HARALICK} feature vectors extracted from the selected segments, (f) displays the feature vectors separated into three sets (training, validation, and test) according to their regions for use in training the SVM classifier~\cite{svm} during the evolutionary process of the UMDA algorithm~\cite{muhlenbein1996recombination}, (g) exhibits the best individual found by the UMDA algorithm (bands B4, B6, B1, and B7), (h) is an illustrative example of an image composed by the best band combination. Finally, in (i), the image segmented by the DeepLabv3$+$ architecture~\cite{chen2018encoderdecoder} is shown, where white pixels correspond to deforested regions in the input image.}
    \label{fig:workflow}
\end{figure*}

Figure~\ref{fig:workflow} presents an overview of the proposed framework.

\subsection{Creation of Segment Dataset}
\label{sec:dataset}

The dataset created for this study comprises nine distinct Landsat-8 satellite images covering the Xingu River Basin\footnote{Xingu River Basin - \url{https://maps.app.goo.gl/n9cR7s5ZNVtYrSqS7}} region, totaling over 8,514 $km^2$ of the Brazilian Legal Amazon. The experimental protocol employed in this study follows the same procedure as in the work~\cite{foresteyes2019}. Therefore, each of the nine images, composed of seven spectral bands, was subjected to a dimensionality reduction process via the PCA algorithm, reducing from seven to only three bands (the first three principal components). Additionally, a non-supervised segmentation algorithm, SLIC~\cite{SLIC}, was applied to create $23,538$ segments, of which $22,587$ were labeled as 'Forest' and $951$ as 'Non-forest'.
 Upon generation of the segments, a selection phase ensued to identify the optimal segments. Thus, this selection process was based on two criteria: (1) segments exhibiting a Homogeneity Rate (proportion of the majority class) $\geq 0.70$, in accordance with the ground truth provided by PRODES~\cite{prodes2022}; and (2) segments with a minimum area of $70$ pixels. 
 
Table~\ref{tab:datasettotal} provides a description of the nine regions with different dimensions and their representations in km$^2$. The values for 'Forest' and 'Non-forest' are associated with the number of segments identified by SLIC for each class.
    \begin{table}[ht!]
    \centering
    \caption{Landsat-8 Imagery segment dataset created in this work.} 
    \resizebox{8.5cm}{!}{
    \begin{tabular}{|c|c|c|c|c|c|}
    \hline
        \textbf{Region} & \textbf{km$^2$} &\textbf{Dimension} &\textbf{Forest} & \textbf{Non-forest} & \textbf{Total} \\ \hline
        {1} & 930.40 &1,230$\times$ 843 & 2,919  & 224 & 3,143  \\ \hline
        {2} & 1,124.59 & 1,343$\times$ 933 & 2,784  & 359  & 3,143  \\ \hline
        {3} & 875.36 & 977$\times$ 998 & 2,282  & 55  & 2,337  \\ \hline
        {4} & 607.33 & 768$\times$ 879 & 623  & 15  & 638  \\ \hline
        {5} & 694.77 & 928$\times$ 833 & 1,415  & 27  & 1,442  \\ \hline
        {6} & 1,528.65 & 1,788$\times$ 950 & 6,040  & 42  & 6,082  \\ \hline
        {7} & 778.84 & 853$\times$ 1,017 & 1,608  & 73  & 1,681  \\ \hline
        {8} & 925.91 & 971$\times$ 1,064 & 2,971  & 48  & 3,019  \\ \hline
        { 9} & 1,048.45 & 1,047$\times$ 1,115 & 1,945  & 108  & 2,053  \\ \hline
        \hline
         \textbf{Total} &$\sim$\textbf{8,514}& -- & \textbf{22,587} &  \textbf{951} &  \textbf{23,538}\\\hline
    \end{tabular}    }
    \label{tab:datasettotal}
\end{table}

\subsection{Selecting Landsat-8 Bands based on UMDA}
For each of the $23,538$ segments, their respective Haralick texture descriptors~\cite{HARALICK} are computed, comprising $13$ coefficients calculated in four directions for each of the seven spectral bands, resulting in seven feature vectors for each segment. The segments were categorized into three sets (train, validation, and test) based on their respective regions: regions $3$ and $4$ were designated for testing, region $8$ constituted the validation set, while the remaining regions formed the training set.

In our implementation of the UMDA, each individual in the population consists of 7 genes representing the spectral bands that compose a Landsat-8 image. The presence of a band in an individual is denoted by the value $1$, while its absence is denoted by $0$. The algorithm starts with a random population, followed by the computation of the fitness function for each individual. The fitness function is the balanced accuracy computed over the SVM classifier outcomes. 

Subsequently, the top $n$ individuals are selected to compose the set of "parents". Marginal probabilities are then calculated based on the relative frequency of each gene present among all parents, estimating its probability distribution. Considering this distribution for each gene, new individuals are generated as ``child". The population of the next generation is composed of the union of the sets of child and parents. The process is repeated until the stopping criteria is met (the number of generations). Among the various configurations tested for the UMDA algorithm to enhance efficiency in the evolutionary process, the chosen parameters were: $10$ individuals in the population, $10$ generations, the number of selected individuals (parents) equal to $5$, and the number of generated child equal to $5$. In this experiment, we did not use marginal probability limits.

\subsection{Semantic Segmentation for Deforestation Detection}

In this stage, the nine different original images in the Landsat-8 database, as described in Table~\ref{tab:datasettotal}, undergo a reconstruction process by composing the most relevant spectral bands (i.e., those with the highest representativity among the best individuals from UMDA, as shown in Table~\ref{tab:best_individuals}). These images are then divided into two sets, with the training set composed of seven regions (1, 2, 5, 6, 7, 8, and 9), and the test set consisting of two regions (3 and 4). Similar to the work in~\cite{andrade2020evaluation}, the manual configuration of the training and test sets was chosen, ensuring that images with clouds were not present in the test set.

During the training process of the DeepLabv3$+$ architecture~\cite{chen2018encoderdecoder}, each image from the training and test sets is cropped into sub-images with dimensions $256 \times 256$. In the training set, the cropping procedure utilized a stride of $64$ pixels, while for the test images, the cropping was performed with a stride of $256$ pixels, implying the absence of overlap between patches. Data augmentation is applied to the training set, which undergoes rotation transformations only. This resulted in $9,432$ images for training and $144$ for testing. In addition to the band combinations found through the EDA, we also evaluated a conventional approach: all the seven bands $+$ NDVI~\cite{torres_2021}. For all combinations, the following training protocol was adopted: the DeepLabv3$+$ model was initialized with its encoder block pretrained on Imagenet~\cite{chen2018encoderdecoder, deng2009imagenet} and trained for 100 epochs on our dataset with the ``Jaccard Loss"~\cite{duqueariasJaccardLoss} loss function and a learning rate of $5 \times 10^{-4}$ with the Adam~\cite{kingma2017adam} optimizer. At the end of each training epoch, comes a validation epoch and the model performance metrics (Accuracy, F1-Score, Intersection over Union (IoU), Precision and Recall) are calculated. For comparison purposes, the metrics from the validation epoch with the highest IoU score were selected for analysis.

\section{Results and Discussion}
\label{sec:results}

In this section, we evaluate the results obtained through the proposed framework. The compositions inferred from the feature ranking showed in Table~\ref{tab:best_individuals} are compared with the baselines in the classification and semantic segmentation tasks.

\subsection{Segment classification through SVM}
In this section, we conduct a comparative analysis of the segment classification results performed by SVM classifier for the deforestation detection task.

Table~\ref{tab:best_individuals} presents the top $22$ individuals derived from the UMDA-based framework, accompanied by their corresponding balanced accuracies in the classification task across the designated test regions. To discern the most relevant bands for the target task, we computed the relative frequency of each band and employed it to ascertain feature ranking~\cite{saeys2004featureRanking}. Notably, the bands B$4$, B$3$, and B$1$ exhibit high frequency, appearing in more than half of the best individuals. This underscores the significant importance of these features in the context of deforestation detection tasks.

Table ~\ref{tab:best_individuals_baselines} shows a comparative analysis between the best individual obtained through the UMDA and three commonly used spectral band combinations for deforestation detection~\cite{torres_2021, foresteyes2019, ortega2019evaluation}. As can be seen, the best individual outperforms all evaluated approaches, with relative gains of $2,86\%$ and $6,10\%$ for PCA~\cite{foresteyes2019} and RGB~\cite{ortega2019evaluation}, respectively, and up to $17.88\%$ for All seven bands with NDVI (All$+$NDVI).


\subsection{Deforestation Detection through Semantic Segmentation}
In order to highlight the importance of the most highest-ranking spectral bands identified in the previous experiment, Tables \ref{tab:x03_metrics} and \ref{tab:x04_metrics} present a comparative analysis of semantic segmentation using the DeepLabv3$+$ architecture with different variations of the most frequent spectral bands for two regions from the test set (3 and 4). 

In all performed experiments, we employed the DeepLabv3$+$ architecture with {encoder} weights pre-trained on the ImageNet database~\cite{deng2009imagenet}. Subsequently, we fine-tuned the model with the following parameters: 100 training epochs, a learning rate of $5 \times 10^{-4}$, and the ``Jaccard Loss" cost function. Additionally, we also performed this experiment using all seven bands plus the NDVI index stacked along as an eighth band, following our baseline approach~\cite{torres_2021}. Although the best band combinations vary among regions, it is worth noting that the best combinations for all of them were obtained through our UMDA-based framework, outperforming the baseline approach (All$+$NDVI). 

\begin{figure}[ht!]
    \centering
    \begin{tabular}{cc}
        \includegraphics[width=0.45\linewidth]{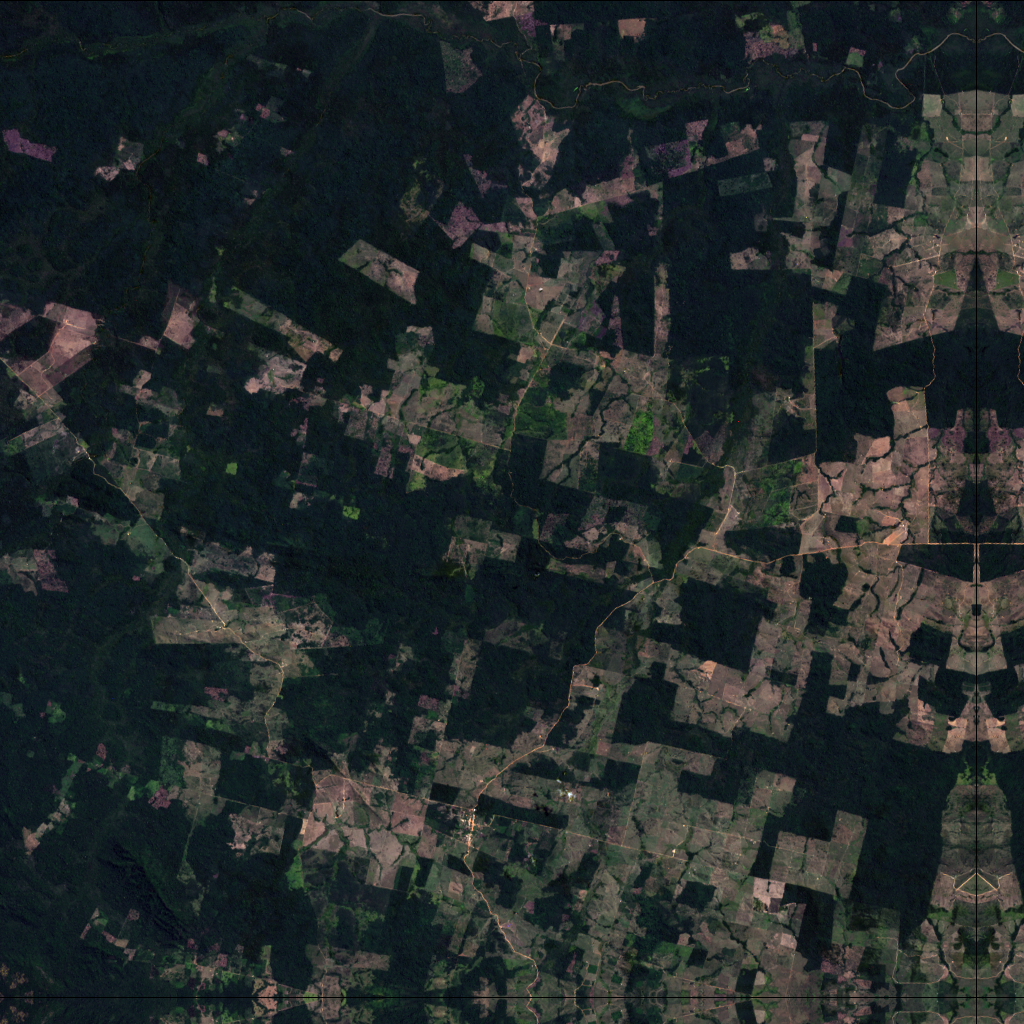}&
        \includegraphics[width=0.45\linewidth]{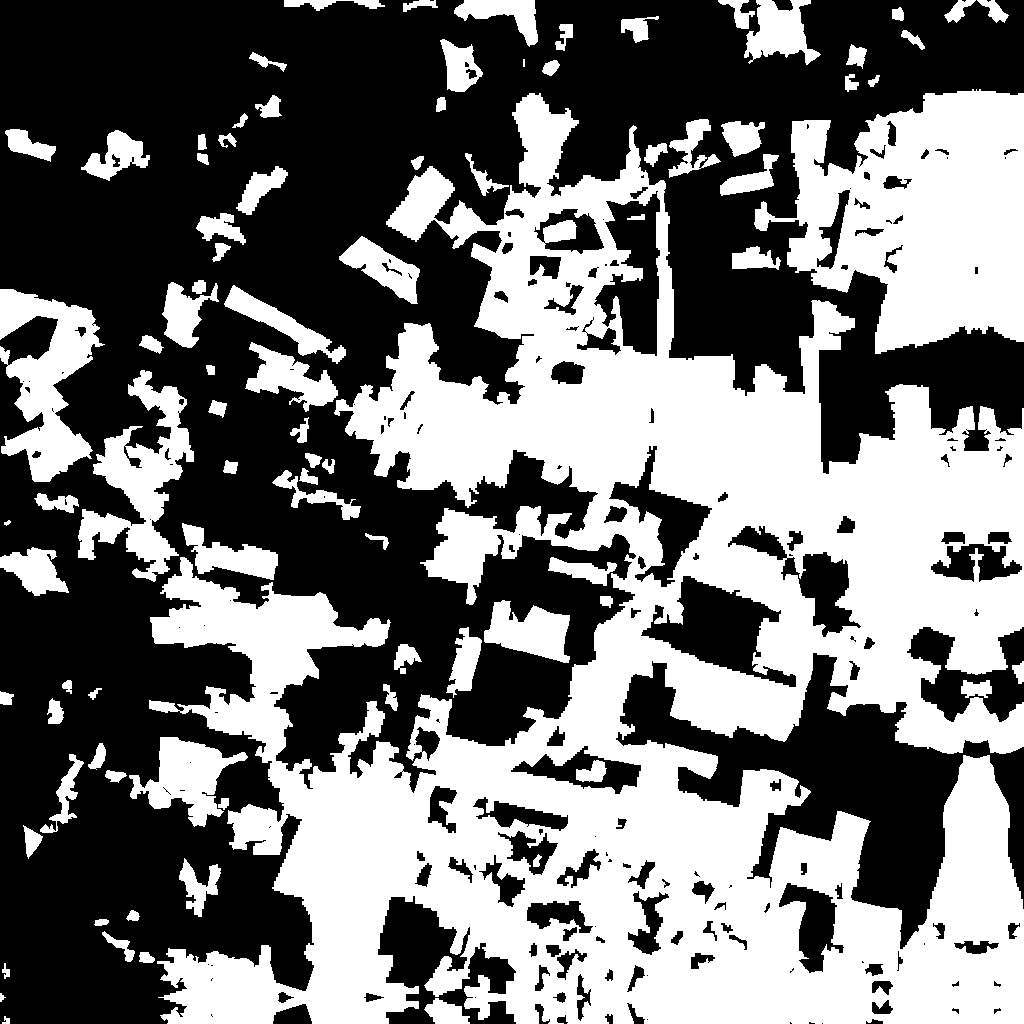} \\
        (a) Region 3 (RGB).&
        (b) Ground truth.\\

        \includegraphics[width=0.45\linewidth]{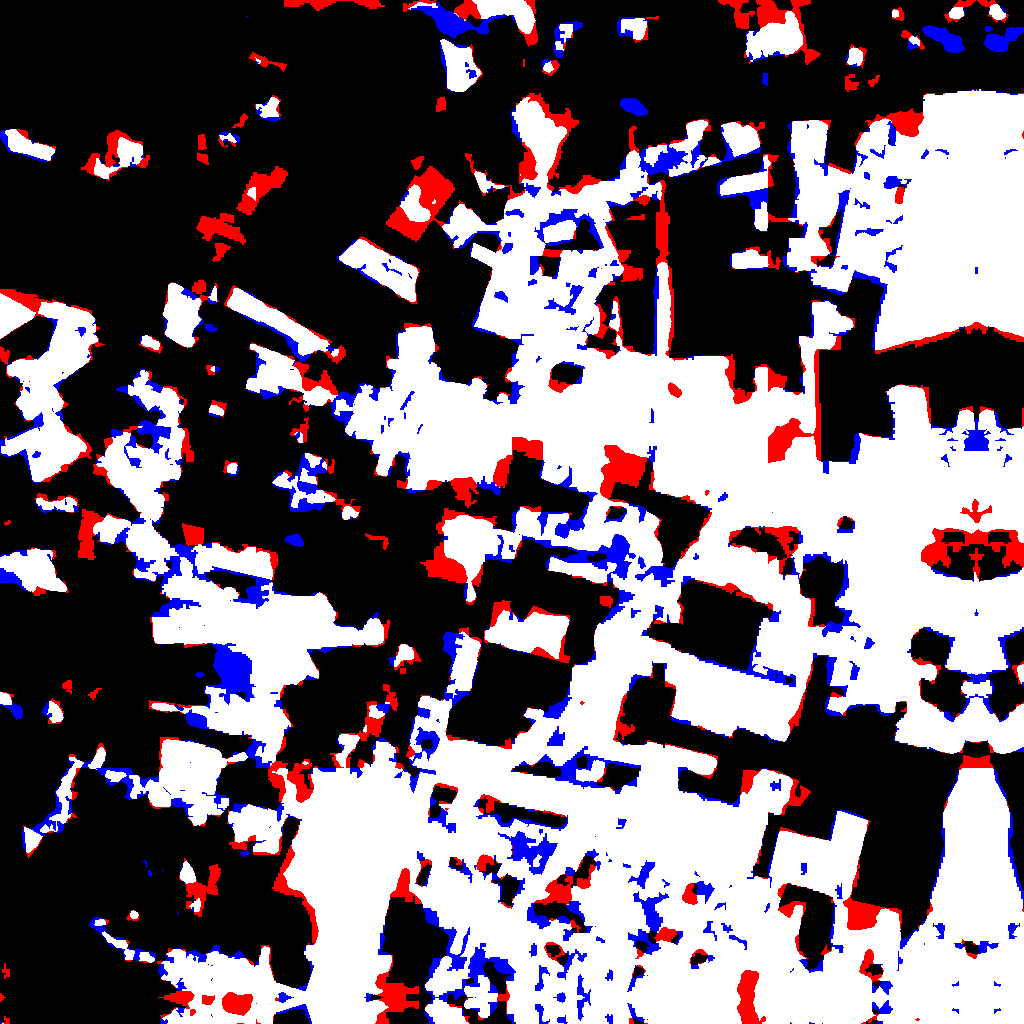}&
        \includegraphics[width=0.45\linewidth]{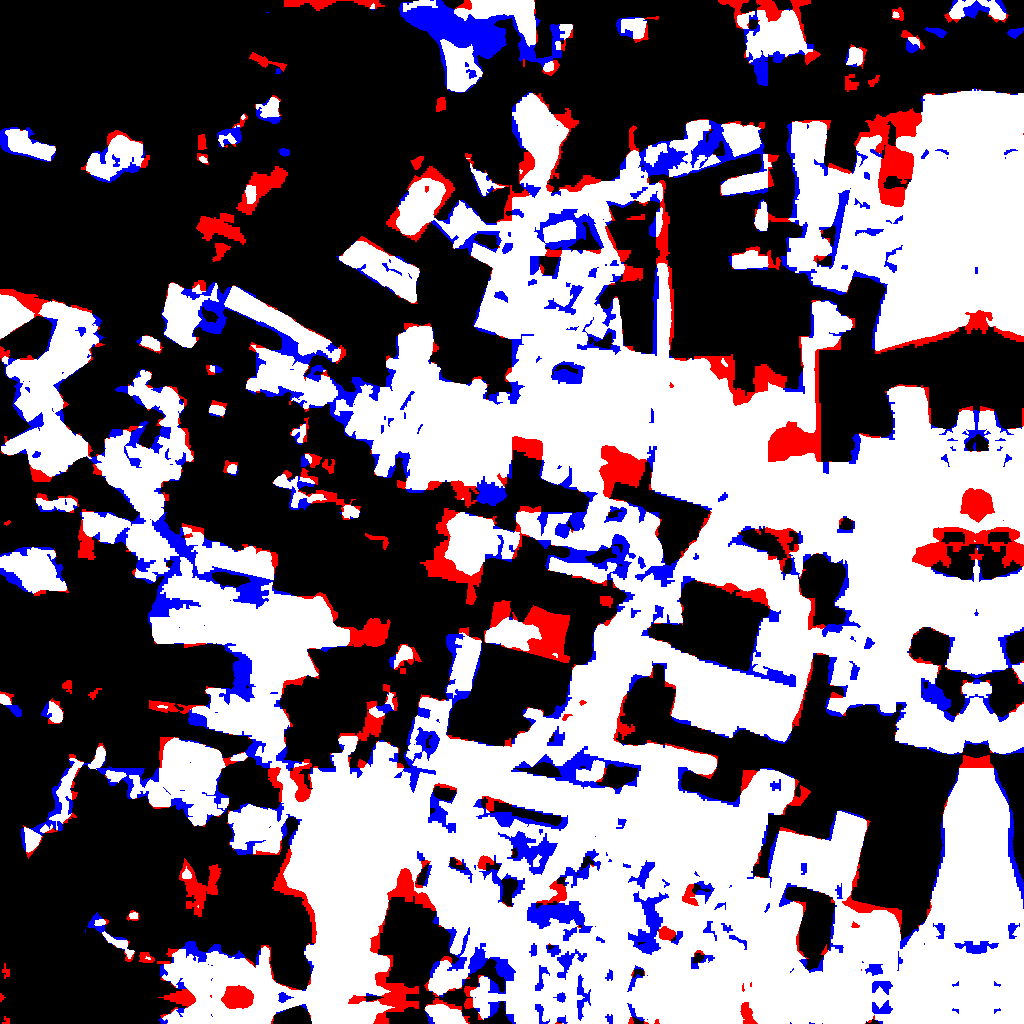}
        \\
        (c) All $+$ NDVI~\cite{torres_2021}.&
        (d) \textbf{(Ours)} B4B3B1B6.\\
 \end{tabular}        
    \caption{In (a) is the region $3$ from the test set composed of the RGB bands. In (b) is the ground truth of the region $3$. In (c) is the semantic segmentation result of the baseline composition from the paper~\cite{torres_2021}. Finally, in (d), is our the best composition band found by UMDA-based framework.}
    \label{fig:x03_results}
\end{figure}

\begin{figure}[ht!]
    \centering
    \begin{tabular}{cc}
        \includegraphics[width=0.45\linewidth]{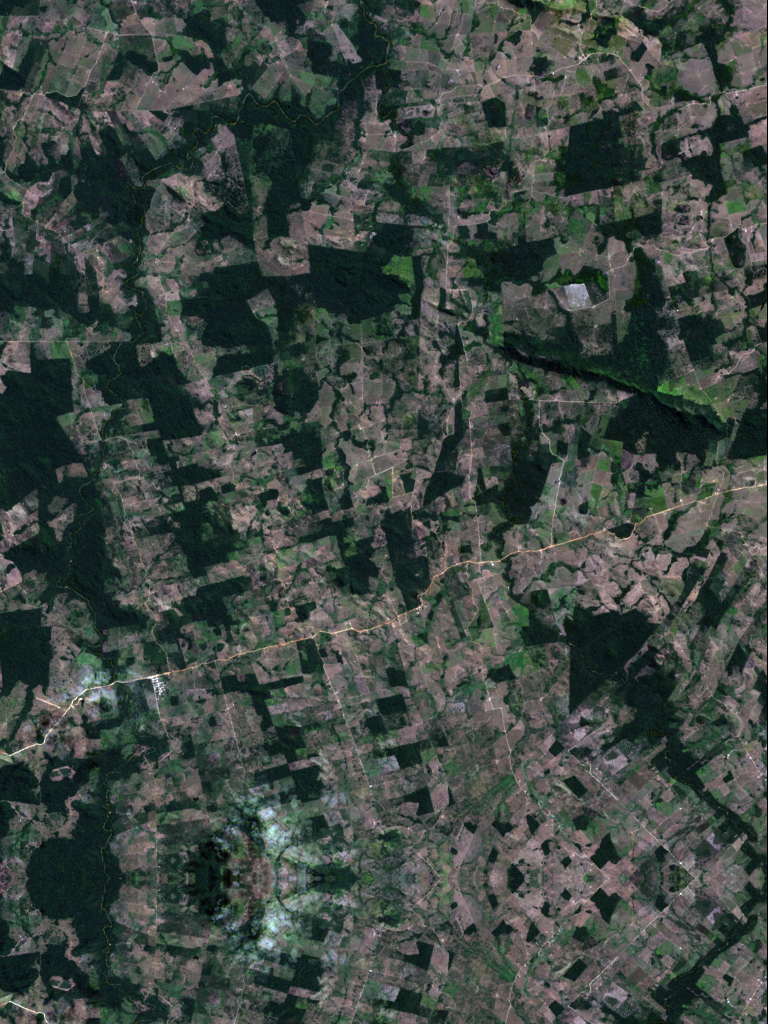}&
        \includegraphics[width=0.45\linewidth]{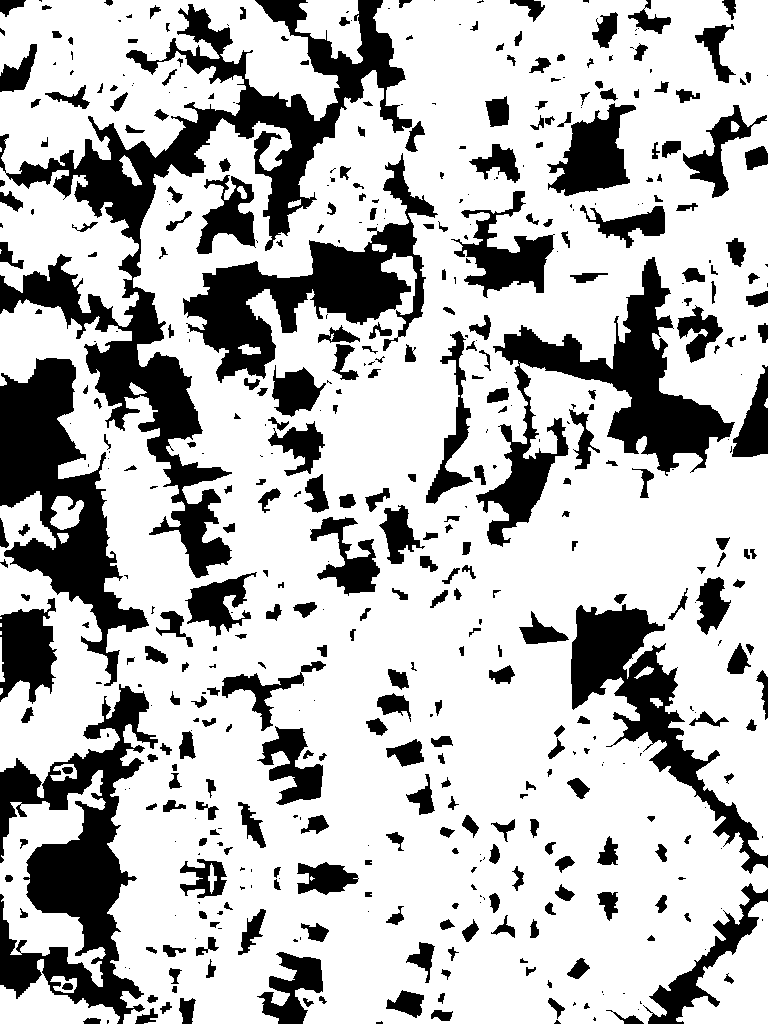}\\
                (a) Region 4 (RGB).&
        (b) Ground truth.\\

\includegraphics[width=0.45\linewidth]{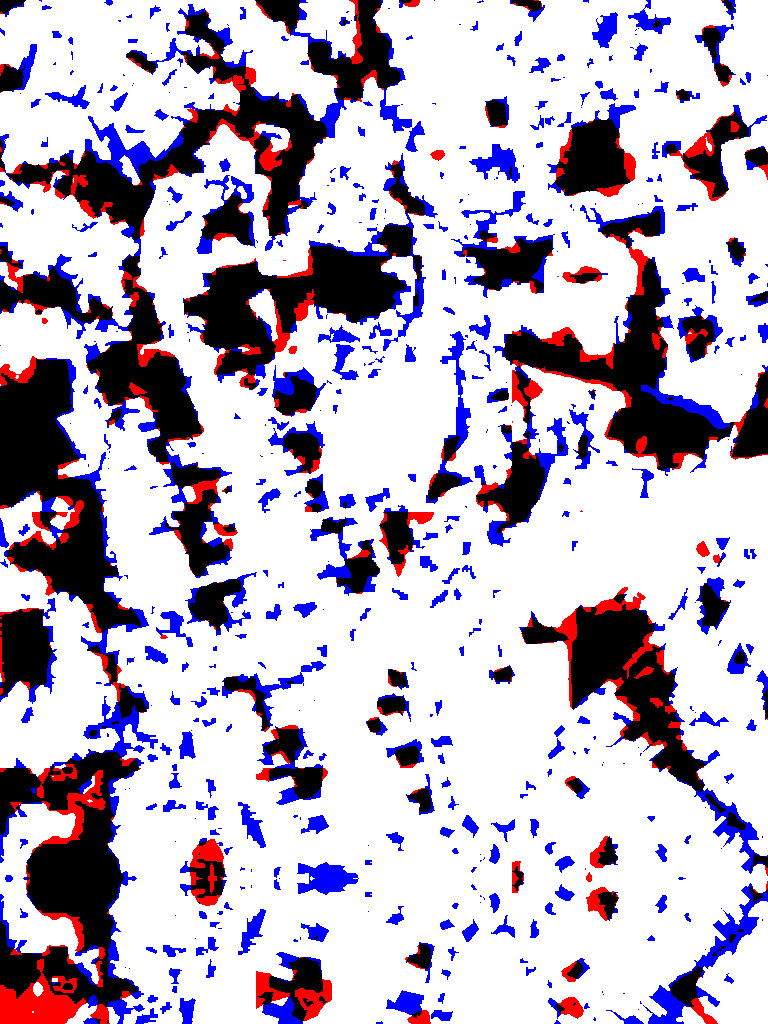}&
        \includegraphics[width=0.45\linewidth]{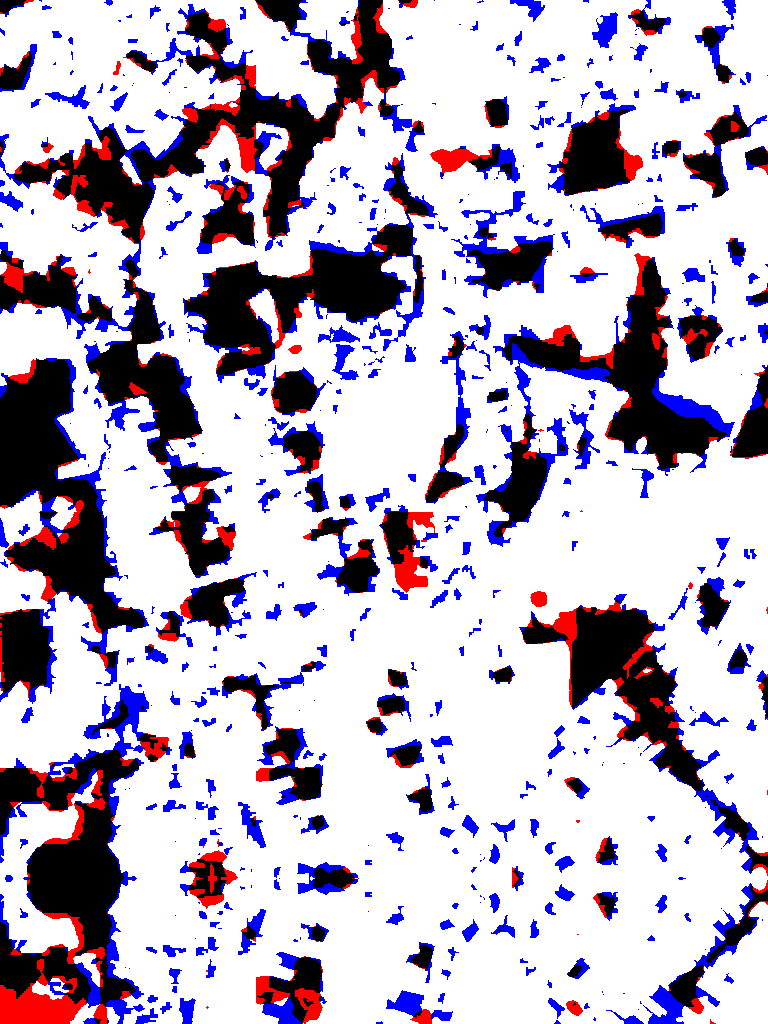}
        \\
        (c) All $+$ NDVI~\cite{torres_2021}.&
        (d) \textbf{(Ours)} B4B3B1B7.\\
 \end{tabular}        
    \caption{In (a) is the region $4$ from the test set composed of the RGB bands. In (b) is the ground truth of the region $4$. In (c) is the semantic segmentation result of the baseline composition from the paper~\cite{torres_2021}. Finally, in (d), is our the best composition band found by UMDA-based framework.}
    \label{fig:x04_results}
\end{figure}

As used in the literature, the criterion for determining the best band compositions was the Intersection over Union (IoU) index. To illustrate the deforestation detection task, the semantic segmentation results for the two regions from the test set (3 and 4) using the All+NDVI and the best-performing combination found by this work are exposed in Figures~\ref{fig:x03_results} and~\ref{fig:x04_results}. In black pixels are True Negative, white pixels are True Positive (non-forest/deforestation), red pixels are False Negative, and blue are True Positive. The original Landsat-8 images in the RGB composition and the PRODES ground truth are also shown.

As observed, the best results in the target task for both regions were achieved by four-bands combinations derived from the proposed framework: B4, B3, B1 and B6 for region $3$ and B4, B3, B1 and B7 for region $4$. This suggests that the spectral band selection framework based on UMDA may bring significant benefits in terms of efficiency (reduced computational cost due to dimensionality reduction) and effectiveness (quality of results) for the semantic segmentation task of deforestation images in the Amazon.

\begin{table*}
    \centering
    \caption{Results of the SVM classifier with the best individuals found by the UMDA algorithm. The symbol \checkmark denotes the presence of the spectral band in the individual.}
    \resizebox{13cm}{!}{
    \begin{tabular}{|c|c|c|c|c|c|c|c|c|c|} \hline
\multirow{2}{*}{\textbf{Seed}} & \textbf{UMDA} & \multicolumn{7}{c|}{\textbf{Bands}} & \textbf{Balanced}\\  \cline{3-9}
 & \textbf{Individual} & \textbf{B1} & \textbf{B2} & \textbf{B3} & \textbf{B4} & \textbf{B5} & \textbf{B6} & \textbf{B7} & \textbf{Accuracy}\\  \hline
 1 & 1 & \checkmark & & \checkmark & \checkmark & & \checkmark & & 87.08\\ \hline
 30 & 2 & \checkmark & & \checkmark & \checkmark & & \checkmark & & 87.08\\ \hline
 1 & 3 & & & & \checkmark & & & & 86.64\\ \hline
 10 & 4 & & & & \checkmark & & & & 86.64\\ \hline
 1 & 5 & \checkmark & & \checkmark & \checkmark & & & & 86.11\\ \hline
 30 & 6 & \checkmark & & \checkmark & \checkmark & & & & 86.11\\ \hline
 42 & 7 & & & \checkmark & & & \checkmark & \checkmark & 84.82\\ \hline
 42 & 8 & \checkmark & & & \checkmark & & & & 84.69\\ \hline
 10 & 9 & \checkmark & & & \checkmark & & & & 84.69\\ \hline
 30 & 10 & & & & \checkmark & & \checkmark & & 84.24\\ \hline
 1 & 11 & & & \checkmark & & & \checkmark & & 83.96\\ \hline
 42 & 12 & & & \checkmark & & & \checkmark & & 83.96\\ \hline
 10 & 13 & \checkmark & & \checkmark & & & & \checkmark & 83.82\\ \hline
 1 & 14 & \checkmark & & \checkmark & \checkmark & & \checkmark & \checkmark & 83.66\\ \hline
 20 & 15 & \checkmark & & \checkmark & \checkmark & & \checkmark & \checkmark & 83.66\\ \hline
 42 & 16 & \checkmark & & \checkmark & \checkmark & & \checkmark & \checkmark & 83.66\\ \hline
 30 & 17 & \checkmark & & \checkmark & \checkmark & & & \checkmark & 83.59\\ \hline
 10 & 18 & \checkmark & & \checkmark & \checkmark & & & \checkmark & 83.59\\ \hline
 1 & 19 & & & \checkmark & \checkmark & & & \checkmark & 83.34\\ \hline
 10 & 20 & & & \checkmark & \checkmark & & & \checkmark & 83.34\\ \hline
 30 & 21 & & & \checkmark & \checkmark & & & \checkmark & 83.34\\ \hline
 20 & 22 & \checkmark & & & & & \checkmark & & 83.17\\ \hline\hline
  
\textbf{Frequency} & - & 59.1\%	& 0.00\% & 72.7\% & 72.7\%	& 0.00 \% &45.5\% & 45.5\%& -\\  \hline 
\textbf{Ranking} & - & \textbf{3$^{\circ}$} & – & \textbf{1$^{\circ}$} & \textbf{1$^{\circ}$} & – & \textbf{4$^{\circ}$} & \textbf{4$^{\circ}$} & -\\  \hline
\end{tabular}
}
\label{tab:best_individuals}
\end{table*}

\begin{table}[ht!]
    \centering
    \caption{Comparison of the SVM classification results among best UMDA individual and three different compositions commonly used in deforestation detection tasks.}
    {
    \begin{tabular}{|c|c|c|} \hline
    \textbf{Composition} & \textbf{Balanced Accuracy} & \textbf{Relative Gain}\\ \hline
    {PCA~\cite{foresteyes2019}} & $84.66\%$ & $2.86\%$\\ \hline
    {RGB~\cite{ortega2019evaluation}} & $82.07\%$ & $6.10\%$\\ \hline
    {All$+$NDVI} & $73.87\%$ & $17.88\%$\\ \hline 
    \best{Best UMDA Ind.} & \best{87.08\%} &  \textbf{--} \\ \hline
\end{tabular}
    }
\label{tab:best_individuals_baselines}
\end{table}

\begin{table}[!ht]
    \centering
    \caption{Semantic segmentation results for the Region 3  using DeepLabv3$+$, sorted by IoU. The best values for each metric are highlighted.}
    \begin{tabular}{|c|c|c|c|c|c|}
    \hline
        \textbf{Composition} & \textbf{Precision} & \textbf{Recall} & \textbf{F1} & \textbf{Acc} & \textbf{IoU} \\ \hline
        B4B3B1B6 & 83.68 & 85.93 & \best{84.60} & 89.33 & \best{74.28} \\ \hline
        All+NDVI~\cite{torres_2021} & \best{86.46} & 83.26 & 84.24 & \best{89.97} & 74.17 \\ \hline
        B4B3B1B7 & 83.12 & 86.47 & 84.45 & 89.17 & 74.09 \\ \hline
        B4B3B1 & 82.97 & 85.71 & 83.82 & 89.45 & 73.63 \\ \hline
        B4B3B1B6B7 & 79.38 & \best{88.67} & 83.41 & 88.40 & 72.83 \\ \hline
        B4 & 81.35 & 85.64 & 83.20 & 88.51 & 72.50 \\ \hline
        B4B3 & 83.17 & 82.75 & 82.67 & 88.37 & 71.76 \\ \hline
    \end{tabular}
    
    \label{tab:x03_metrics}
\end{table}

\begin{table}[!ht]
    \centering
     \caption{Semantic segmentation results for the Region 4 using DeepLabv3$+$, sorted by IoU. The best values for each metric are highlighted.}
    \begin{tabular}{|c|c|c|c|c|c|}
    \hline
        \textbf{Composition} & \textbf{Precision} & \textbf{Recall} & \textbf{F1} & \textbf{Acc} & \textbf{IoU} \\ \hline
        B4B3B1B7 & \best{89.65} & 95.43 & \best{92.43} & \best{88.93} & \best{86.01} \\ \hline
        B4 & 88.71 & 96.22 & 92.30 & 88.64 & 85.78 \\ \hline
        All+NDVI~\cite{torres_2021} & 89.33 & 95.06 & 92.07 & 88.37 & 85.38 \\ \hline
        B4B3B1B6 & 87.20 & 96.87 & 91.76 & 87.65 & 84.86 \\ \hline
        B4B3B1B6B7 & 85.82 & 97.81 & 91.39 & 86.88 & 84.24 \\ \hline
        B4B3 & 86.32 & 96.65 & 91.16 & 86.68 & 83.86 \\ \hline
        B4B3B1 & 84.69 & \best{98.16} & 90.89 & 85.98 & 83.40 \\ \hline
    \end{tabular}
   
    \label{tab:x04_metrics}
\end{table}

\section{Conclusions}
\label{sec:conclusao}
In this study, a novel framework has been proposed for the detection of deforested areas in the Amazon. Through the selection of Landsat-8 satellite bands based on the stochastic evolutionary algorithm UMDA, it was possible to identify the bands that contribute most significantly to the representation of the areas of interest. Armed with this knowledge, images from the dataset were generated using only the most relevant spectral bands, feeding into a semantic segmentation architecture (DeepLabv3$+$) for the deforestation detection task. The conducted experiments successfully demonstrated that the proposed spectral band selection framework outperforms commonly employed approaches in the literature, achieving greater performance for deforestation detection through classification and semantic segmentation. {Interestingly, this framework also surpasses the performance approaches that utilize a larger number of bands. This finding challenges the widely accepted data-driven principle in deep learning, reinforcing the notion that ‘more’ does not necessarily mean ‘better’.}
Despite the specific combination of both different areas, it is crucial to identify an optimal band set to facilitate generalization for semantic segmentation, a task intended for future handling.

\section*{Acknowledgements}

The authors would like to express their gratitude to the funding agency CNPq for providing the PIBIC scholarship to student Eduardo B. Neto. 
This research is part of the National Institute of Science and Technology (INCT) of the Future Internet for Smart Cities, funded by CNPq (grant number 465446/2014-0), CAPES (Finance Code 001), and FAPESP (grants numbers 2014/50937-1, 2015/24485-9, 2017/25908-6, 2018/23908-1, 2019/26702-8, 2023/00811-0, and 2023/00782-0). 
The authors acknowledge the National Laboratory for Scientific Computing (LNCC/MCTI, Brazil) for providing HPC resources of the SDumont supercomputer, which have contributed to the research results reported within this paper\footnote{SDumont - \url{http://sdumont.lncc.br}}.
\clearpage
\balance
\bibliographystyle{ACM-Reference-Format}
\bibliography{sample-base}


\begin{thebibliography}{41}


\ifx \showCODEN    \undefined \def \showCODEN     #1{\unskip}     \fi
\ifx \showDOI      \undefined \def \showDOI       #1{#1}\fi
\ifx \showISBNx    \undefined \def \showISBNx     #1{\unskip}     \fi
\ifx \showISBNxiii \undefined \def \showISBNxiii  #1{\unskip}     \fi
\ifx \showISSN     \undefined \def \showISSN      #1{\unskip}     \fi
\ifx \showLCCN     \undefined \def \showLCCN      #1{\unskip}     \fi
\ifx \shownote     \undefined \def \shownote      #1{#1}          \fi
\ifx \showarticletitle \undefined \def \showarticletitle #1{#1}   \fi
\ifx \showURL      \undefined \def \showURL       {\relax}        \fi
\providecommand\bibfield[2]{#2}
\providecommand\bibinfo[2]{#2}
\providecommand\natexlab[1]{#1}
\providecommand\showeprint[2][]{arXiv:#2}

\bibitem[\protect\citeauthoryear{Achanta, Shaji, Smith, Lucchi, Fua, and S{\"u}sstrunk}{Achanta et~al\mbox{.}}{2012}]%
        {SLIC}
\bibfield{author}{\bibinfo{person}{Radhakrishna Achanta}, \bibinfo{person}{Appu Shaji}, \bibinfo{person}{Kevin Smith}, \bibinfo{person}{Aurelien Lucchi}, \bibinfo{person}{Pascal Fua}, {and} \bibinfo{person}{Sabine S{\"u}sstrunk}.} \bibinfo{year}{2012}\natexlab{}.
\newblock \showarticletitle{S{LIC} superpixels compared to state-of-the-art superpixel methods}.
\newblock \bibinfo{journal}{\emph{IEEE transactions on pattern analysis and machine intelligence}} \bibinfo{volume}{34}, \bibinfo{number}{11} (\bibinfo{year}{2012}), \bibinfo{pages}{2274--2282}.
\newblock


\bibitem[\protect\citeauthoryear{Alexandre}{Alexandre}{2017}]%
        {alexandre2017ift}
\bibfield{author}{\bibinfo{person}{Eduardo~Barreto Alexandre}.} \bibinfo{year}{2017}\natexlab{}.
\newblock \emph{\bibinfo{title}{IFT-SLIC: gera{\c{c}}{\~a}o de superpixels com base em agrupamento iterativo linear simples e transformada imagem-floresta}}.
\newblock \bibinfo{thesistype}{Master's\ thesis}. \bibinfo{school}{Universidade de S{\~a}o Paulo}.
\newblock


\bibitem[\protect\citeauthoryear{Amroabadi, Ahmadzadeh, and Hekmatnia}{Amroabadi et~al\mbox{.}}{2011}]%
        {amroabadi2011haralickmammograms}
\bibfield{author}{\bibinfo{person}{SayedMasoud~Hashemi Amroabadi}, \bibinfo{person}{Mohammad~Reza Ahmadzadeh}, {and} \bibinfo{person}{Ali Hekmatnia}.} \bibinfo{year}{2011}\natexlab{}.
\newblock \showarticletitle{Mass detection in mammograms using ga based PCA and Haralick features selection}. In \bibinfo{booktitle}{\emph{2011 19th Iranian Conference on Electrical Engineering}}. \bibinfo{pages}{1--4}.
\newblock


\bibitem[\protect\citeauthoryear{Andrade, Costa, Mota, Ortega, Feitosa, Soto, and Heipke}{Andrade et~al\mbox{.}}{2020}]%
        {andrade2020evaluation}
\bibfield{author}{\bibinfo{person}{RB Andrade}, \bibinfo{person}{GAOP Costa}, \bibinfo{person}{GLA Mota}, \bibinfo{person}{MX Ortega}, \bibinfo{person}{RQ Feitosa}, \bibinfo{person}{PJ Soto}, {and} \bibinfo{person}{Christian Heipke}.} \bibinfo{year}{2020}\natexlab{}.
\newblock \showarticletitle{Evaluation of semantic segmentation methods for deforestation detection in the amazon}.
\newblock \bibinfo{journal}{\emph{ISPRS Archives; 43, B3}} \bibinfo{volume}{43}, \bibinfo{number}{B3} (\bibinfo{year}{2020}), \bibinfo{pages}{1497--1505}.
\newblock


\bibitem[\protect\citeauthoryear{Boser, Guyon, and Vapnik}{Boser et~al\mbox{.}}{1992}]%
        {svm}
\bibfield{author}{\bibinfo{person}{B.~E. Boser}, \bibinfo{person}{I.~M. Guyon}, {and} \bibinfo{person}{V.~N. Vapnik}.} \bibinfo{year}{1992}\natexlab{}.
\newblock \showarticletitle{A training algorithm for optimal margin classifiers}. In \bibinfo{booktitle}{\emph{Workshop on Computational Learning Theory}} (Pittsburgh, Pennsylvania, United States) \emph{(\bibinfo{series}{COLT '92})}. \bibinfo{pages}{144--152}.
\newblock
\showISBNx{0-89791-497-X}


\bibitem[\protect\citeauthoryear{Chen, Lu, Zhu, and Zhang}{Chen et~al\mbox{.}}{2023}]%
        {chen2023generative}
\bibfield{author}{\bibinfo{person}{Jiaqi Chen}, \bibinfo{person}{Jiachen Lu}, \bibinfo{person}{Xiatian Zhu}, {and} \bibinfo{person}{Li Zhang}.} \bibinfo{year}{2023}\natexlab{}.
\newblock \bibinfo{title}{Generative Semantic Segmentation}.
\newblock
\newblock
\showeprint[arxiv]{2303.11316}~[cs.CV]


\bibitem[\protect\citeauthoryear{Chen, Zhu, Papandreou, Schroff, and Adam}{Chen et~al\mbox{.}}{2018}]%
        {chen2018encoderdecoder}
\bibfield{author}{\bibinfo{person}{Liang-Chieh Chen}, \bibinfo{person}{Yukun Zhu}, \bibinfo{person}{George Papandreou}, \bibinfo{person}{Florian Schroff}, {and} \bibinfo{person}{Hartwig Adam}.} \bibinfo{year}{2018}\natexlab{}.
\newblock \bibinfo{title}{Encoder-Decoder with Atrous Separable Convolution for Semantic Image Segmentation}.
\newblock
\newblock
\showeprint[arxiv]{1802.02611}~[cs.CV]


\bibitem[\protect\citeauthoryear{Dallaqua, Faria, and Álvaro Fazenda}{Dallaqua et~al\mbox{.}}{2021}]%
        {dallaquaSIBGRAPI}
\bibfield{author}{\bibinfo{person}{Fernanda Dallaqua}, \bibinfo{person}{Fabio Faria}, {and} \bibinfo{person}{Álvaro Fazenda}.} \bibinfo{year}{2021}\natexlab{}.
\newblock \showarticletitle{ForestEyes Project - Citizen Science and Machine Learning to detect deforested areas in tropical forests}. In \bibinfo{booktitle}{\emph{Anais Estendidos do XXXIV Conference on Graphics, Patterns and Images}} (Online). \bibinfo{publisher}{SBC}, \bibinfo{address}{Porto Alegre, RS, Brasil}, \bibinfo{pages}{14--20}.
\newblock
\showISSN{0000-0000}
\urldef\tempurl%
\url{https://doi.org/10.5753/sibgrapi.est.2021.20008}
\showDOI{\tempurl}


\bibitem[\protect\citeauthoryear{Dallaqua, Fazenda, and Faria}{Dallaqua et~al\mbox{.}}{2019}]%
        {foresteyes2019}
\bibfield{author}{\bibinfo{person}{F.B.J.R. Dallaqua}, \bibinfo{person}{A.L. Fazenda}, {and} \bibinfo{person}{F.A. Faria}.} \bibinfo{year}{2019}\natexlab{}.
\newblock \showarticletitle{{ForestEyes} {P}roject: {C}an {C}itizen {S}cientists {H}elp {R}ainforests?}. In \bibinfo{booktitle}{\emph{{IEEE} 15th {International} {Conference} on {eScience}}}. \bibinfo{publisher}{{IEEE}}, \bibinfo{pages}{18--27}.
\newblock


\bibitem[\protect\citeauthoryear{de~Assis, Ferreira, Vinhas, Maurano, de~Almeida, Nascimento, de~Carvalho, Camargo, and Maciel}{de~Assis et~al\mbox{.}}{2019}]%
        {deterrabrasilis}
\bibfield{author}{\bibinfo{person}{Luiz Fernando Ferreira~Gomes de Assis}, \bibinfo{person}{Karine~Reis Ferreira}, \bibinfo{person}{L{\'u}bia Vinhas}, \bibinfo{person}{Luis Maurano}, \bibinfo{person}{Cl{\'a}udio~Aparecido de Almeida}, \bibinfo{person}{Jether~Rodrigues Nascimento}, \bibinfo{person}{Andr{\'e} Fernandes~Ara{\'u}jo de Carvalho}, \bibinfo{person}{Claudinei Camargo}, {and} \bibinfo{person}{Adeline~Marinho Maciel}.} \bibinfo{year}{2019}\natexlab{}.
\newblock \showarticletitle{TERRA{B}RASILIS: A SPATIAL DATA INFRASTRUCTURE FOR DISSEMINATING DEFORESTATION DATA FROM {B}RAZIL}.
\newblock \bibinfo{journal}{\emph{Proceedings of the XIX Brazilian Symposium on Remote Sensing, Santos, Brazil}} (\bibinfo{year}{2019}), \bibinfo{pages}{14--17}.
\newblock


\bibitem[\protect\citeauthoryear{Deng, Dong, Socher, Li, Li, and Fei-Fei}{Deng et~al\mbox{.}}{2009}]%
        {deng2009imagenet}
\bibfield{author}{\bibinfo{person}{Jia Deng}, \bibinfo{person}{Wei Dong}, \bibinfo{person}{Richard Socher}, \bibinfo{person}{Li-Jia Li}, \bibinfo{person}{Kai Li}, {and} \bibinfo{person}{Li Fei-Fei}.} \bibinfo{year}{2009}\natexlab{}.
\newblock \showarticletitle{Imagenet: A large-scale hierarchical image database}. In \bibinfo{booktitle}{\emph{2009 IEEE conference on computer vision and pattern recognition}}. Ieee, \bibinfo{pages}{248--255}.
\newblock


\bibitem[\protect\citeauthoryear{Duque-Arias, Velasco-Forero, Deschaud, Goulette, Serna, Decenci{\`e}re, and Marcotegui}{Duque-Arias et~al\mbox{.}}{2021}]%
        {duqueariasJaccardLoss}
\bibfield{author}{\bibinfo{person}{David Duque-Arias}, \bibinfo{person}{Santiago Velasco-Forero}, \bibinfo{person}{Jean-Emmanuel Deschaud}, \bibinfo{person}{Francois Goulette}, \bibinfo{person}{Andr{\'e}s Serna}, \bibinfo{person}{Etienne Decenci{\`e}re}, {and} \bibinfo{person}{Beatriz Marcotegui}.} \bibinfo{year}{2021}\natexlab{}.
\newblock \showarticletitle{{On power Jaccard losses for semantic segmentation}}. In \bibinfo{booktitle}{\emph{{VISAPP 2021 : 16th International Conference on Computer Vision Theory and Applications}}}. \bibinfo{address}{Vienne (on line), Austria}.
\newblock
\urldef\tempurl%
\url{https://hal.science/hal-03139997}
\showURL{%
\tempurl}


\bibitem[\protect\citeauthoryear{{et al.}}{{et al.}}{2023}]%
        {doi:10.1126/science.abp8622}
\bibfield{author}{\bibinfo{person}{David M.~Lapola {et al.}}} \bibinfo{year}{2023}\natexlab{}.
\newblock \showarticletitle{The drivers and impacts of Amazon forest degradation}.
\newblock \bibinfo{journal}{\emph{Science}} \bibinfo{volume}{379}, \bibinfo{number}{6630} (\bibinfo{year}{2023}), \bibinfo{pages}{eabp8622}.
\newblock
\urldef\tempurl%
\url{https://doi.org/10.1126/science.abp8622}
\showDOI{\tempurl}
\showeprint{https://www.science.org/doi/pdf/10.1126/science.abp8622}


\bibitem[\protect\citeauthoryear{Fernandez-Moral, Martins, Wolf, and Rives}{Fernandez-Moral et~al\mbox{.}}{2018}]%
        {Fernandez2018SSmetrics}
\bibfield{author}{\bibinfo{person}{Eduardo Fernandez-Moral}, \bibinfo{person}{Renato Martins}, \bibinfo{person}{Denis Wolf}, {and} \bibinfo{person}{Patrick Rives}.} \bibinfo{year}{2018}\natexlab{}.
\newblock \showarticletitle{A New Metric for Evaluating Semantic Segmentation: Leveraging Global and Contour Accuracy}. In \bibinfo{booktitle}{\emph{2018 IEEE Intelligent Vehicles Symposium (IV)}}. \bibinfo{pages}{1051--1056}.
\newblock
\urldef\tempurl%
\url{https://doi.org/10.1109/IVS.2018.8500497}
\showDOI{\tempurl}


\bibitem[\protect\citeauthoryear{Hansen, Potapov, Moore, Hancher, Turubanova, Tyukavina, Thau, Stehman, Goetz, Loveland, et~al\mbox{.}}{Hansen et~al\mbox{.}}{2013}]%
        {Hansen850}
\bibfield{author}{\bibinfo{person}{Matthew~C Hansen}, \bibinfo{person}{Peter~V Potapov}, \bibinfo{person}{Rebecca Moore}, \bibinfo{person}{Matt Hancher}, \bibinfo{person}{SAA Turubanova}, \bibinfo{person}{Alexandra Tyukavina}, \bibinfo{person}{David Thau}, \bibinfo{person}{SV Stehman}, \bibinfo{person}{SJ Goetz}, \bibinfo{person}{Thomas~R Loveland}, {et~al\mbox{.}}} \bibinfo{year}{2013}\natexlab{}.
\newblock \showarticletitle{High-resolution global maps of 21st-century forest cover change}.
\newblock \bibinfo{journal}{\emph{science}} \bibinfo{volume}{342}, \bibinfo{number}{6160} (\bibinfo{year}{2013}), \bibinfo{pages}{850--853}.
\newblock


\bibitem[\protect\citeauthoryear{Haralick, Shanmugam, et~al\mbox{.}}{Haralick et~al\mbox{.}}{1973}]%
        {HARALICK}
\bibfield{author}{\bibinfo{person}{Robert~M Haralick}, \bibinfo{person}{Karthikeyan Shanmugam}, {et~al\mbox{.}}} \bibinfo{year}{1973}\natexlab{}.
\newblock \showarticletitle{Textural features for image classification}.
\newblock \bibinfo{journal}{\emph{IEEE Transactions on systems, man, and cybernetics}} \bibinfo{number}{6} (\bibinfo{year}{1973}), \bibinfo{pages}{610--621}.
\newblock


\bibitem[\protect\citeauthoryear{Hauschild and Pelikan}{Hauschild and Pelikan}{2011}]%
        {HAUSCHILD2011111}
\bibfield{author}{\bibinfo{person}{Mark Hauschild} {and} \bibinfo{person}{Martin Pelikan}.} \bibinfo{year}{2011}\natexlab{}.
\newblock \showarticletitle{{An introduction and survey of estimation of distribution algorithms}}.
\newblock \bibinfo{journal}{\emph{Swarm and Evolutionary Computation}} \bibinfo{volume}{1}, \bibinfo{number}{3} (\bibinfo{year}{2011}), \bibinfo{pages}{111--128}.
\newblock


\bibitem[\protect\citeauthoryear{INPE}{INPE}{2022}]%
        {prodes2022}
\bibfield{author}{\bibinfo{person}{INPE}.} \bibinfo{year}{2022}\natexlab{}.
\newblock \bibinfo{title}{Estimativa de desmatamento na {A}mazônia {L}egal para $2022$ é de $11.568$ $km^2$}.
\newblock \bibinfo{howpublished}{https://encurtador.com.br/aBDP2}.
\newblock
\newblock
\shownote{Accessed: 2023-08-01}.


\bibitem[\protect\citeauthoryear{Irons, Dwyer, and Barsi}{Irons et~al\mbox{.}}{2012}]%
        {landsat8Specs}
\bibfield{author}{\bibinfo{person}{James~R. Irons}, \bibinfo{person}{John~L. Dwyer}, {and} \bibinfo{person}{Julia~A. Barsi}.} \bibinfo{year}{2012}\natexlab{}.
\newblock \showarticletitle{The next Landsat satellite: The Landsat Data Continuity Mission}.
\newblock \bibinfo{journal}{\emph{Remote Sensing of Environment}}  \bibinfo{volume}{122} (\bibinfo{year}{2012}), \bibinfo{pages}{11--21}.
\newblock
\showISSN{0034-4257}
\urldef\tempurl%
\url{https://doi.org/10.1016/j.rse.2011.08.026}
\showDOI{\tempurl}
\newblock
\shownote{Landsat Legacy Special Issue}.


\bibitem[\protect\citeauthoryear{Jolliffe}{Jolliffe}{2011}]%
        {PCA}
\bibfield{author}{\bibinfo{person}{Ian Jolliffe}.} \bibinfo{year}{2011}\natexlab{}.
\newblock \bibinfo{booktitle}{\emph{Principal component analysis}}.
\newblock \bibinfo{publisher}{Springer}.
\newblock


\bibitem[\protect\citeauthoryear{Kingma and Ba}{Kingma and Ba}{2017}]%
        {kingma2017adam}
\bibfield{author}{\bibinfo{person}{Diederik~P. Kingma} {and} \bibinfo{person}{Jimmy Ba}.} \bibinfo{year}{2017}\natexlab{}.
\newblock \bibinfo{title}{Adam: A Method for Stochastic Optimization}.
\newblock
\newblock
\showeprint[arxiv]{1412.6980}~[cs.LG]


\bibitem[\protect\citeauthoryear{Landsat {S}cience}{Landsat {S}cience}{2017}]%
        {landsat8Overview}
Landsat {S}cience \bibinfo{year}{2017}\natexlab{}.
\newblock \bibinfo{title}{Landsat {S}cience}.
\newblock \bibinfo{howpublished}{\url{https://landsat.gsfc.nasa.gov/}}.
\newblock
\newblock
\shownote{Accessed:2017-12-15}.


\bibitem[\protect\citeauthoryear{Long, Shelhamer, and Darrell}{Long et~al\mbox{.}}{2015}]%
        {long2015fully}
\bibfield{author}{\bibinfo{person}{Jonathan Long}, \bibinfo{person}{Evan Shelhamer}, {and} \bibinfo{person}{Trevor Darrell}.} \bibinfo{year}{2015}\natexlab{}.
\newblock \showarticletitle{Fully convolutional networks for semantic segmentation}. In \bibinfo{booktitle}{\emph{Proceedings of the IEEE conference on computer vision and pattern recognition}}. \bibinfo{pages}{3431--3440}.
\newblock


\bibitem[\protect\citeauthoryear{Luz et~al\mbox{.}}{Luz et~al\mbox{.}}{2014}]%
        {AFW14}
\bibfield{author}{\bibinfo{person}{Eduardo~F.P. Luz} {et~al\mbox{.}}} \bibinfo{year}{2014}\natexlab{}.
\newblock \showarticletitle{The {F}orest{W}atchers: {A} {C}itizen {C}yberscience {P}roject for {D}eforestation {M}onitoring in the {T}ropics}.
\newblock \bibinfo{journal}{\emph{Human Computation}}  \bibinfo{volume}{1} (\bibinfo{year}{2014}), \bibinfo{pages}{137--145}.
\newblock


\bibitem[\protect\citeauthoryear{Ma, Zheng, Tong, and Zheng}{Ma et~al\mbox{.}}{2003}]%
        {ma2003bandselection}
\bibfield{author}{\bibinfo{person}{Ji-Ping Ma}, \bibinfo{person}{Zhao-Bao Zheng}, \bibinfo{person}{Qing-Xi Tong}, {and} \bibinfo{person}{Lan-Fen Zheng}.} \bibinfo{year}{2003}\natexlab{}.
\newblock \showarticletitle{An application of genetic algorithms on band selection for hyperspectral image classification}. In \bibinfo{booktitle}{\emph{Proceedings of the 2003 international conference on machine learning and cybernetics (IEEE Cat. No. 03EX693)}}, Vol.~\bibinfo{volume}{5}. IEEE, \bibinfo{pages}{2810--2813}.
\newblock


\bibitem[\protect\citeauthoryear{Maciel and Vinhas}{Maciel and Vinhas}{2019}]%
        {macielreasoning}
\bibfield{author}{\bibinfo{person}{Adeline~Marinho Maciel} {and} \bibinfo{person}{Lubia Vinhas}.} \bibinfo{year}{2019}\natexlab{}.
\newblock \showarticletitle{Reasoning about deforestation trajectories in {P}ar\'a state, Brazil}. In \bibinfo{booktitle}{\emph{Anais do XIX Simpósio Brasileiro de Sensoriamento Remoto}}.
\newblock


\bibitem[\protect\citeauthoryear{Maretto, Fonseca, Jacobs, K{\"o}rting, Bendini, and Parente}{Maretto et~al\mbox{.}}{2020}]%
        {maretto2020spatio}
\bibfield{author}{\bibinfo{person}{Raian~V Maretto}, \bibinfo{person}{Leila~MG Fonseca}, \bibinfo{person}{Nathan Jacobs}, \bibinfo{person}{Thales~S K{\"o}rting}, \bibinfo{person}{Hugo~N Bendini}, {and} \bibinfo{person}{Leandro~L Parente}.} \bibinfo{year}{2020}\natexlab{}.
\newblock \showarticletitle{Spatio-temporal deep learning approach to map deforestation in amazon rainforest}.
\newblock \bibinfo{journal}{\emph{IEEE Geoscience and Remote Sensing Letters}} \bibinfo{volume}{18}, \bibinfo{number}{5} (\bibinfo{year}{2020}), \bibinfo{pages}{771--775}.
\newblock


\bibitem[\protect\citeauthoryear{Martin}{Martin}{2015}]%
        {martin2015edge}
\bibfield{author}{\bibinfo{person}{Claude Martin}.} \bibinfo{year}{2015}\natexlab{}.
\newblock \bibinfo{booktitle}{\emph{On the {E}dge: {T}he {S}tate and {F}ate of the {W}orld's {T}ropical {R}ainforests}}.
\newblock \bibinfo{publisher}{Greystone Books Ltd}.
\newblock


\bibitem[\protect\citeauthoryear{M{\"u}hlenbein and Paass}{M{\"u}hlenbein and Paass}{1996}]%
        {muhlenbein1996recombination}
\bibfield{author}{\bibinfo{person}{Heinz M{\"u}hlenbein} {and} \bibinfo{person}{Gerhard Paass}.} \bibinfo{year}{1996}\natexlab{}.
\newblock \showarticletitle{From recombination of genes to the estimation of distributions I. Binary parameters}. In \bibinfo{booktitle}{\emph{International conference on parallel problem solving from nature}}. Springer, \bibinfo{pages}{178--187}.
\newblock


\bibitem[\protect\citeauthoryear{Nagasubramanian, Jones, Sarkar, Singh, Singh, and Ganapathysubramanian}{Nagasubramanian et~al\mbox{.}}{2018}]%
        {nagasubramanian2018hyperspectral_ga}
\bibfield{author}{\bibinfo{person}{Koushik Nagasubramanian}, \bibinfo{person}{Sarah Jones}, \bibinfo{person}{Soumik Sarkar}, \bibinfo{person}{Asheesh~K Singh}, \bibinfo{person}{Arti Singh}, {and} \bibinfo{person}{Baskar Ganapathysubramanian}.} \bibinfo{year}{2018}\natexlab{}.
\newblock \showarticletitle{Hyperspectral band selection using genetic algorithm and support vector machines for early identification of charcoal rot disease in soybean stems}.
\newblock \bibinfo{journal}{\emph{Plant methods}}  \bibinfo{volume}{14} (\bibinfo{year}{2018}), \bibinfo{pages}{1--13}.
\newblock


\bibitem[\protect\citeauthoryear{Nowozin}{Nowozin}{2014}]%
        {Nowozin_2014_CVPR}
\bibfield{author}{\bibinfo{person}{Sebastian Nowozin}.} \bibinfo{year}{2014}\natexlab{}.
\newblock \showarticletitle{Optimal Decisions from Probabilistic Models: The Intersection-over-Union Case}. In \bibinfo{booktitle}{\emph{Proceedings of the IEEE Conference on Computer Vision and Pattern Recognition (CVPR)}}.
\newblock


\bibitem[\protect\citeauthoryear{Ortega, Bermudez, Happ, Gomes, and Feitosa}{Ortega et~al\mbox{.}}{2019}]%
        {ortega2019evaluation}
\bibfield{author}{\bibinfo{person}{MX Ortega}, \bibinfo{person}{JD Bermudez}, \bibinfo{person}{PN Happ}, \bibinfo{person}{A Gomes}, {and} \bibinfo{person}{RQ Feitosa}.} \bibinfo{year}{2019}\natexlab{}.
\newblock \showarticletitle{Evaluation of deep learning techniques for deforestation detection in the Amazon forest}.
\newblock \bibinfo{journal}{\emph{ISPRS Annals of the Photogrammetry, Remote Sensing and Spatial Information Sciences}}  \bibinfo{volume}{4} (\bibinfo{year}{2019}), \bibinfo{pages}{121--128}.
\newblock


\bibitem[\protect\citeauthoryear{Saeys, Degroeve, Aeyels, Rouz{\'e}, and Van~de Peer}{Saeys et~al\mbox{.}}{2004}]%
        {saeys2004featureRanking}
\bibfield{author}{\bibinfo{person}{Yvan Saeys}, \bibinfo{person}{Sven Degroeve}, \bibinfo{person}{Dirk Aeyels}, \bibinfo{person}{Pierre Rouz{\'e}}, {and} \bibinfo{person}{Yves Van~de Peer}.} \bibinfo{year}{2004}\natexlab{}.
\newblock \showarticletitle{Feature selection for splice site prediction: a new method using EDA-based feature ranking}.
\newblock \bibinfo{journal}{\emph{BMC bioinformatics}}  \bibinfo{volume}{5} (\bibinfo{year}{2004}), \bibinfo{pages}{1--11}.
\newblock


\bibitem[\protect\citeauthoryear{Souza, Vieira~Monteiro, Daleles~Rennó, Almeida, de~Morisson~Valeriano, Morelli, Vinhas, P.~Maurano, Adami, Sobral~Escada, da~Motta, and Amaral}{Souza et~al\mbox{.}}{2019}]%
        {metodprodes2019}
\bibfield{author}{\bibinfo{person}{Arlesson Souza}, \bibinfo{person}{Antônio~Miguel Vieira~Monteiro}, \bibinfo{person}{Camilo Daleles~Rennó}, \bibinfo{person}{Cláudio~Aparecido Almeida}, \bibinfo{person}{Dalton de Morisson~Valeriano}, \bibinfo{person}{Fabiano Morelli}, \bibinfo{person}{Lubia Vinhas}, \bibinfo{person}{Luis~Eduardo P.~Maurano}, \bibinfo{person}{Marcos Adami}, \bibinfo{person}{Maria~Isabel Sobral~Escada}, \bibinfo{person}{Marisa da Motta}, {and} \bibinfo{person}{Silvana Amaral}.} \bibinfo{year}{2019}\natexlab{}.
\newblock \showarticletitle{Metodologia {U}tilizada nos {P}rojetos {P}{R}{O}{D}{E}{S} e {D}{E}{T}{E}{R}}.
\newblock \bibinfo{journal}{\emph{S{\~a}o Jos{\'e} dos Campos: INPE}} (\bibinfo{year}{2019}).
\newblock


\bibitem[\protect\citeauthoryear{Tian, Zhao, Son, Luo, Oh, and Wang}{Tian et~al\mbox{.}}{2023}]%
        {tian2023AtmosphericRivers}
\bibfield{author}{\bibinfo{person}{Yuan Tian}, \bibinfo{person}{Yang Zhao}, \bibinfo{person}{Seok-Woo Son}, \bibinfo{person}{Jing-Jia Luo}, \bibinfo{person}{Seok-Geun Oh}, {and} \bibinfo{person}{Yinjun Wang}.} \bibinfo{year}{2023}\natexlab{}.
\newblock \showarticletitle{A Deep-Learning Ensemble Method to Detect Atmospheric Rivers and Its Application to Projected Changes in Precipitation Regime}.
\newblock \bibinfo{journal}{\emph{Journal of Geophysical Research: Atmospheres}} \bibinfo{volume}{128}, \bibinfo{number}{12} (\bibinfo{year}{2023}), \bibinfo{pages}{e2022JD037041}.
\newblock
\urldef\tempurl%
\url{https://doi.org/10.1029/2022JD037041}
\showDOI{\tempurl}
\showeprint{https://agupubs.onlinelibrary.wiley.com/doi/pdf/10.1029/2022JD037041}
\newblock
\shownote{e2022JD037041 2022JD037041}.


\bibitem[\protect\citeauthoryear{Torres, Turnes, Soto~Vega, Feitosa, Silva, Marcato~Junior, and Almeida}{Torres et~al\mbox{.}}{2021}]%
        {torres_2021}
\bibfield{author}{\bibinfo{person}{Daliana~Lobo Torres}, \bibinfo{person}{Javier~Noa Turnes}, \bibinfo{person}{Pedro~Juan Soto~Vega}, \bibinfo{person}{Raul~Queiroz Feitosa}, \bibinfo{person}{Daniel~E. Silva}, \bibinfo{person}{Jose Marcato~Junior}, {and} \bibinfo{person}{Claudio Almeida}.} \bibinfo{year}{2021}\natexlab{}.
\newblock \showarticletitle{Deforestation Detection with Fully Convolutional Networks in the Amazon Forest from Landsat-8 and Sentinel-2 Images}.
\newblock \bibinfo{journal}{\emph{Remote Sensing}} \bibinfo{volume}{13}, \bibinfo{number}{24} (\bibinfo{year}{2021}).
\newblock


\bibitem[\protect\citeauthoryear{Urquhart, Chomentowski, Skole, and Barber}{Urquhart et~al\mbox{.}}{1998}]%
        {nasaflorestas}
\bibfield{author}{\bibinfo{person}{Gerald Urquhart}, \bibinfo{person}{Walter Chomentowski}, \bibinfo{person}{David Skole}, {and} \bibinfo{person}{Chris Barber}.} \bibinfo{year}{1998}\natexlab{}.
\newblock \bibinfo{title}{Tropical deforestation}.
\newblock
\newblock


\bibitem[\protect\citeauthoryear{Walker, Stickler, Kellndorfer, Kirsch, and Nepstad}{Walker et~al\mbox{.}}{2010}]%
        {PRODES3}
\bibfield{author}{\bibinfo{person}{Wayne~S Walker}, \bibinfo{person}{Claudia~M Stickler}, \bibinfo{person}{Josef~M Kellndorfer}, \bibinfo{person}{Katie~M Kirsch}, {and} \bibinfo{person}{Daniel~C Nepstad}.} \bibinfo{year}{2010}\natexlab{}.
\newblock \showarticletitle{Large-area classification and mapping of forest and land cover in the Brazilian Amazon: A comparative analysis of {ALOS}/{PALSAR} and {L}andsat data sources}.
\newblock \bibinfo{journal}{\emph{IEEE Journal of selected topics in applied earth observations and remote sensing}} \bibinfo{volume}{3}, \bibinfo{number}{4} (\bibinfo{year}{2010}), \bibinfo{pages}{594--604}.
\newblock


\bibitem[\protect\citeauthoryear{Wei, Zhang, Liu, and Zhao}{Wei et~al\mbox{.}}{1705}]%
        {WeiFeatureSelection2017}
\bibfield{author}{\bibinfo{person}{Bin Wei}, \bibinfo{person}{Minqing Zhang}, \bibinfo{person}{Longfei Liu}, {and} \bibinfo{person}{Jing Zhao}.} \bibinfo{year}{2017/05}\natexlab{}.
\newblock \showarticletitle{Feature Selection on the Basis of Rough Set Theory and Univariate Marginal Distribution Algorithm}. In \bibinfo{booktitle}{\emph{Proceedings of the 2017 International Conference on Applied Mathematics, Modelling and Statistics Application (AMMSA 2017)}}. \bibinfo{publisher}{Atlantis Press}, \bibinfo{pages}{369--372}.
\newblock
\showISBNx{978-94-6252-355-5}
\showISSN{1951-6851}
\urldef\tempurl%
\url{https://doi.org/10.2991/ammsa-17.2017.83}
\showDOI{\tempurl}


\bibitem[\protect\citeauthoryear{Yu, Di, Yang, Tang, Lin, Zhang, Rahman, Zhao, Gaigalas, Yu, et~al\mbox{.}}{Yu et~al\mbox{.}}{2019}]%
        {yu2019selection}
\bibfield{author}{\bibinfo{person}{Zhiqi Yu}, \bibinfo{person}{Liping Di}, \bibinfo{person}{Ruixing Yang}, \bibinfo{person}{Junmei Tang}, \bibinfo{person}{Li Lin}, \bibinfo{person}{Chen Zhang}, \bibinfo{person}{Md~Shahinoor Rahman}, \bibinfo{person}{Haoteng Zhao}, \bibinfo{person}{Juozas Gaigalas}, \bibinfo{person}{Eugene~Genong Yu}, {et~al\mbox{.}}} \bibinfo{year}{2019}\natexlab{}.
\newblock \showarticletitle{Selection of landsat 8 OLI band combinations for land use and land cover classification}. In \bibinfo{booktitle}{\emph{2019 8th International Conference on Agro-Geoinformatics (Agro-Geoinformatics)}}. IEEE, \bibinfo{pages}{1--5}.
\newblock


\bibitem[\protect\citeauthoryear{Zhang, Sun, and Li}{Zhang et~al\mbox{.}}{2009}]%
        {zhang2009bandselection}
\bibfield{author}{\bibinfo{person}{Xianfeng Zhang}, \bibinfo{person}{Quan Sun}, {and} \bibinfo{person}{Jonathan Li}.} \bibinfo{year}{2009}\natexlab{}.
\newblock \showarticletitle{Optimal band selection for high dimensional remote sensing data using genetic algorithm}. In \bibinfo{booktitle}{\emph{Second International Conference on Earth Observation for Global Changes}}, Vol.~\bibinfo{volume}{7471}. SPIE, \bibinfo{pages}{522--528}.
\newblock


\end{thebibliography}

\end{document}